\documentclass{article} 
\usepackage{arxiv}

\usepackage{natbib}
\usepackage{hyperref}
\usepackage{url}
\usepackage{amssymb}
\usepackage{amsthm}
\usepackage{graphicx}

\usepackage{booktabs}
\usepackage{multirow}
\usepackage{array}
\usepackage{tabularx}
\usepackage{amsmath}

\newtheorem{definition}{\textbf{Definition}}[section]

\title{Cross-Embodiment Causal Transfer from Human Videos to Robot Manipulation}

\author{
Xingyu Zhang$^{1,2}$,
Yingchao Li$^{3}$,
Peizheng Guo$^{2,1}$,
Jie Hu$^{3}$,
Changwen Zheng$^{2}$,
Wenwen Qiang$^{2}$\thanks{Corresponding author.}
\\[0.8em]
$^{1}$University of Chinese Academy of Sciences\\
$^{2}$Institute of Software, Chinese Academy of Sciences\\
$^{3}$Meituan\\[0.5em]
\texttt{Correspondence: qiangwenwen@iscas.ac.cn}
}

\begin{document}

\maketitle

\begin{abstract}
Scaling robot manipulation models requires large amounts of interaction data, yet collecting diverse real-robot experience remains costly. Human manipulation videos, in contrast, provide abundant and diverse task experience that could potentially benefit robot learning. However, differences in morphology, action spaces, and control mechanisms make human motions unsuitable as direct supervision for robot actions, raising a scientific question: what knowledge in human videos should be transferred across embodiments to robot manipulation? To address this question, we analyze subtask transfer through the lens of causal beams, focusing on which contextual conditions and interaction dependencies should be retained to support local task effects. Based on this perspective, we propose Causal-Beam-Guided Subtask Transfer (CBST). CBST learns manipulation-related representations from human videos and grounds them in task-level semantic variables using robot demonstrations. Within a predefined manipulation model, causal-beam constraints guide contextual information selection by requiring retained assignments to sustain local mechanism outputs while preserving nontrivial responsiveness to retained inputs. The selected semantic information conditions a world-action model, allowing the robot to generate actions from its own observations without requiring human–robot motion correspondence. Experiments on robot manipulation benchmarks show improvements in task success over the corresponding world-action model baselines.
\end{abstract}



\section{Introduction}
Recent robot manipulation models are evolving toward World Action Models (WAMs) that jointly model future scene evolution and robot behavior, enabling predictions of interaction outcomes \citep{cosmospolicy2026,lingbotva2026}. However, scaling real-world robot data requires dedicated hardware, human operation, and physical deployment, making broad coverage of tasks, objects, and scenes expensive. In contrast, human manipulation videos are abundant and capture diverse task processes and object interactions, making them an attractive source of additional experience for robot learning \citep{droid2024,humanvideosurvey2026,mimicplay2023}. Yet humans and robots differ substantially in morphology, action spaces, and control mechanisms, so concrete human actions cannot be directly reused as robot supervision \citep{whirl2022,xskill2023,uniskill2025}. This leads to the central question of this work: \textbf{\emph{what knowledge in human manipulation videos should be transferred across embodiment differences to robot manipulation?}}

\begin{figure}
    \centering
    \includegraphics[width=1\linewidth]{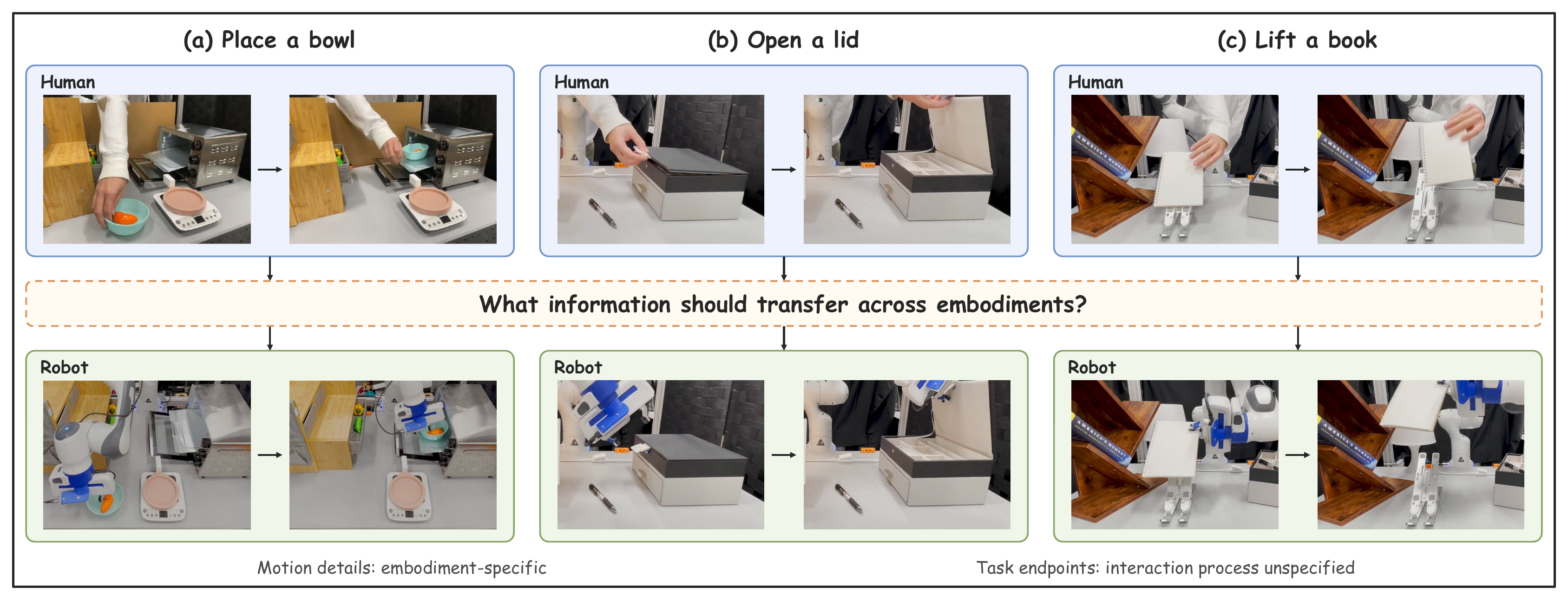}
    \caption{The cross-embodiment transfer problem. Human and robot demonstrations share task objectives but differ in execution. Examples of placing a bowl, opening a lid, and lifting a book motivate the question: what task information should transfer beyond embodiment-specific motion and goal endpoints?}
    \label{fig:placeholder}
\end{figure}

A complete human manipulation sequence contains two types of information: object-interaction relations that support task progress, and concrete motion realizations that are closely tied to the human embodiment. For long-horizon manipulation tasks, we first organize continuous execution into local task processes using temporally extended subtasks, thereby avoiding the requirement that the robot reproduce human motion step by step. However, defining a subtask only specifies the local scope within which transfer takes place; it does not further determine which information within that scope should be preserved together with the task experience. Retaining only the task intention or boundary states of a subtask may discard the conditions and interaction relations that determine how a local effect is realized, whereas retaining the complete human execution trajectory reintroduces substantial embodiment-specific motion details. 

We study this selective information-preservation problem through the lens of causal beams \citep{pearl2009causality}. Given a structural model and a specific context, a causal beam allows part of the inputs to be projected out in a context-dependent manner, while requiring the actual values of the retained variables to sustain the actual outcome of the current mechanism and the projected mechanism to retain nontrivial responsiveness to its retained inputs. This perspective motivates separating task-level abstraction from contextual information selection. Task-level variables provide an abstraction of execution, while causal-beam conditions constrain which semantic dependencies are retained within a specified model.

Based on this, we propose {Causal-Beam-Guided Subtask Transfer (CBST)}. CBST first learns manipulation-related temporal representations through human-video pretraining and grounds them, under robot demonstration supervision, into a candidate SCM instantiated from shared manipulation templates. A contextual subtask-conditioning module then selects a contextual causal beam from the candidate SCM according to the current execution context, producing the local task effect, relevant conditions, and their effect-relevant dependencies. This structural information conditions the target robot WAM, while concrete actions are still generated from the robot's current state and its own action space. CBST therefore combines human-video representation transfer with causal-beam-constrained selection of contextual subtask information for robot execution.

Our contributions are summarized as: (\textbf{1}) We study human-to-robot learning from the perspective of cross-embodiment transfer, focusing on which task-progress structures should remain transferable after embodiment-specific motion is discarded; (\textbf{2}) We propose \textbf{Causal-Beam-Guided Subtask Transfer (CBST)}, which combines a shared candidate manipulation SCM with learned contextual causal beams to preserve effect-relevant conditions and dependencies for conditioning robot WAMs; (\textbf{3}) We systematically evaluate CBST on WAMs including LingBot-VA and Cosmos Policy, examining the contributions of human video pretraining, context-specific causal selection, and their effects on robot control.

\section{Related Works}
Human-to-robot transfer aims to exploit abundant human demonstrations while addressing the embodiment gap between human and robot execution. Motion-oriented approaches seek representations that remain useful across embodiments while retaining execution information. WHIRL aligns human and robot videos to infer task intent and initializes robot policy learning from third-person human demonstrations \citep{whirl2022}, while EgoMimic uses egocentric human videos with 3D hand tracking, cross-domain alignment, and joint training on human and robot demonstrations \citep{egomimic2025}. Latent Action Pretraining (LAPA) instead learns discrete latent actions from videos without robot action labels and subsequently maps them to robot actions using limited robot data \citep{lapa2025}. Other approaches move toward higher-level transferable representations. MimicPlay learns latent plans from large-scale human play videos and uses them to guide a low-level robot controller \citep{mimicplay2023}; XSkill discovers cross-embodiment skill prototypes from unlabeled human and robot videos and maps them to robot actions with a conditional policy \citep{xskill2023}; Translating Robot Skills learns unsupervised correspondences between skill sequences across morphologically different agents \citep{translatingskills2022}; and UniSkill learns embodiment-agnostic skill representations that allow skills extracted from human video prompts to guide robot policies trained on robot data \citep{uniskill2025}. More recently, Gen2Act generates human future videos for novel tasks and conditions a robot policy on the generated motion information \citep{gen2act2025}.

Temporal abstraction organizes long-horizon behavior into reusable units beyond primitive actions. The options framework formalizes temporally extended policies through initiation sets, internal policies, and termination conditions \citep{options1999}, while later robotic methods model skill feasibility and composition, such as SayCan \citep{saycan2022} and relational skill-precondition learning. Another line of work represents task progress through intermediate subgoals: Video Language Planning generates hierarchical visual and language plans \citep{vlp2024}, TaKSIE conditions visual subgoal generation on task progress \citep{taksie2025}, and Latent Space Backward Planning recursively constructs intermediate latent goals from the final task goal \citep{lbp2025}. More recent world-model methods integrate such temporal abstraction directly into planning: Hierarchical Planning with Latent World Models learns dynamics at multiple temporal scales and uses high-level latent subgoals to guide low-level planning \citep{hwm2026}, while SAGE similarly exploits temporally extended latent subgoals for action-sequence generation \citep{sage2026}.

World models learn predictive dynamics to support action selection and planning. Early latent-dynamics approaches such as PlaNet perform planning directly in learned latent spaces \citep{planet2019}, while more recent methods extend this idea to visual robot manipulation: DINO-WM predicts future visual features for zero-shot planning \citep{dinowm2025}, V-JEPA~2-AC learns action-conditioned latent dynamics for model-predictive control \citep{vjepa2_2025}, and IRASim models fine-grained action-conditioned visual futures for simulation and planning \citep{irasim2025}. Recent World-Action Models further unify future prediction with action generation. WorldVLA jointly models visual and action tokens in an autoregressive framework \citep{worldvla2025}; Cosmos Policy generates robot actions together with future-state images and value estimates \citep{cosmospolicy2026}; LingBot-VA jointly learns future-frame prediction and policy execution in a shared visual--action latent space \citep{lingbotva2026}; and $\tau_0$-WM further combines video prediction, action generation, and action evaluation within a unified model \citep{tau0wm2026}.

\section{Problem Definition and Analysis}
\subsection{Problem Formulation}
Robot manipulation is a sequential decision-making process: given a task instruction \(g\), the robot generates actions based on the available information and obtains new observations through interaction with the environment. Let \(s_t\) denote the robot observation at time \(t\), comprising visual input and available proprioceptive measurements, and let \(a_t\) denote a robot action. The interaction history is \(h_t=(s_0,a_0,\ldots,a_{t-1},s_t)\). We consider a world-action model (WAM) that jointly models robot actions and future observations conditioned on the interaction history \(h_t\) and task instruction \(g\).

Given \(h_t\) and \(g\), the WAM generates an action chunk and predicts future observations through the joint conditional distribution \(p_\theta(A_t,S_t^+\mid h_t,g)\). Here, \(A_t=(a_t,\ldots,a_{t+H-1})\) denotes an action chunk of length \(H\), and \(S_t^+=(s_{t+\delta})_{\delta\in\mathcal T_H}\) denotes the future-observation sequence at offsets \(\mathcal T_H\subseteq\{1,\ldots,H\}\). After executing the action chunk, the robot updates the interaction history using newly acquired observations, forming a closed-loop control process.

\subsection{Causal Beams}\label{sec:causal_beams}
We use a recursive structural causal model \(M=(U,V,\mathcal F)\), where each non-root variable \(V_i\in V\) is determined by a structural mechanism \(V_i=f_i(\mathrm{PA}_i,U)\). A context \(U=u\) determines the actual values \(V_i(u)\). For the causal-beam construction, the parent set of each mechanism is partitioned as
\begin{equation}
\mathrm{PA}_i=C_i\mathbin{\dot\cup}N_i,
\end{equation}
where \(C_i\) denotes the retained parents and \(N_i\) the remaining parents. The retained parents must sustain the actual output:
\begin{equation}
f_i(C_i(u),n_i,u)=V_i(u),
\qquad
\forall n_i\in\operatorname{Val}(N_i).
\label{eq:beam_sustain}
\end{equation}

In this work, we use the natural-beam specialization of the general causal-beam construction \citep{pearl2009causality}. The remaining inputs are fixed at their actual values, giving the projected mechanism
\begin{equation}
f_i^u(c_i)=f_i(c_i,N_i(u),u).
\label{eq:natural_projection}
\end{equation}
The projection must remain nontrivially responsive to its retained inputs:
\begin{equation}
\exists c_i'\in\operatorname{Val}(C_i)
\quad\text{s.t.}\quad
f_i^u(c_i')\neq V_i(u).
\label{eq:beam_response}
\end{equation}
Thus, the actual assignments of the retained parents sustain the current mechanism’s output, while the projected mechanism is nonconstant in the retained inputs considered jointly. This local responsiveness condition does not establish that every retained parent is individually necessary or that each retained upstream variable has an intervention effect on the downstream effect \(Y\) in the composed model. A collection of such projected mechanisms forms a natural beam only when their retained and projected variables are mutually compatible across mechanisms. The full compatibility conditions are given in Appendix \ref{app:causal}.

\begin{figure}
    \centering
    \includegraphics[width=1\linewidth]{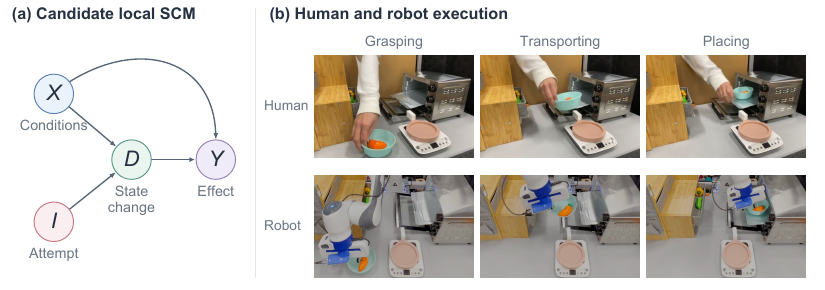}
    \caption{(a) Candidate local manipulation SCM for a fixed task goal, showing the endogenous parent structure. \(X\) and \(I\) are boundary inputs, and the mechanisms for \(D\) and \(Y\) may differ across embodiments. Variable groups are instantiated into task-specific variables before beam selection. 
(b) Illustrative human and robot motion realizations of the same bowl-placement task.}
    \label{fig:local_manipulation_scm}
\end{figure}

\subsection{Causal Analysis of Cross-Embodiment Transfer}
\label{sec:causal_transfer_analysis}
Humans and robots may accomplish the same task through different motion realizations because of differences in morphology, action spaces, and control mechanisms. This motivates describing local task processes using semantic variables for task conditions, manipulation attempts, object-state changes, and effects, without requiring step-by-step correspondence between human and robot motion. Defining these variables, however, does not determine which conditions and dependencies should be retained in a particular execution context. We use causal-beam conditions to guide this contextual selection within a specified task-level structural model.

To formalize the task-relevant structure under consideration, we use a candidate SCM as Fig.\ref{fig:local_manipulation_scm}(a) for a local manipulation process. We assume that human and robot executions of the same local manipulation can be described using a shared set of task-level variable semantics and candidate dependency relations, while their embodiment-specific structural functions may differ. This shared schema defines the hypothesis space in which cross-embodiment structural correspondence is considered; it does not assert that every candidate dependency is active in every execution context. Sharing this schema does not guarantee that a beam valid in one embodiment remains valid in another; the beam conditions must be evaluated relative to the instantiated mechanisms and execution context. Let \(X\) denote task-relevant conditions at the beginning of the interval, \(I\) the manipulation attempt, \(D\) the resulting object-state change, and \(Y\) the realized local effect:
\begin{equation}
X=U_X, I=U_I, D=f_D^{e}(X,I,U_D),Y=f_Y^{e}(X,D,U_Y),
\label{eq:local_manipulation_scm}
\end{equation}
where \(e\in\{h,r\}\) denotes the embodiment. Thus, humans and robots share the task-level variables and the hypothesis space of possible dependencies, but not necessarily the concrete mechanisms that realize these dependencies. The mechanisms \(f_D^{e}\) and \(f_Y^{e}\) may therefore differ between humans and robots. The distinction between \(I\) and \(D\) separates an attempted manipulation from the object change that is actually realized. This schema specifies candidate task-relevant variable roles and dependencies rather than a complete physical model; task-specific variables are instantiated from the corresponding manipulation specification. Within this candidate structure, causal-beam selection determines which dependencies are retained in a particular execution context.

Consider moving a bowl into a drawer. Let \(B\) indicate whether a transport motion is executed, \(G\) whether the grasp is effective, \(O\) whether the drawer entrance is accessible, \(D\) whether the bowl is carried by the transport motion, and \(Y\) whether it reaches the target region. Here \(X=(G,O)\) and \(I=B\). A simplified local model is 
\begin{equation}
    D=B\land G, Y=D\land O.
\end{equation}
In a successful context with \(B=G=O=1\), no proper subset of \(\{B,G\}\) sustains \(D=1\), and no proper subset of \(\{D,O\}\) sustains \(Y=1\). A projection of the original mechanism that omits one of these parents may still match the observed endpoint, but it is not a valid sustaining projection in this context. In contrast, when transport is attempted with an ineffective grasp, \(B=1,G=0,O=1\), one valid natural projection retains \(G\) for \(D\) and \(D\) for \(Y\), yielding \(D=G\) and \(Y=D\). Thus, the same candidate SCM can admit different valid projections under different execution contexts. Appendix~\ref{app:causal} provides the full verification of the causal-beam conditions.
This example highlights two points relevant to transfer. First, matching an observed endpoint does not imply that the dependencies producing that endpoint have been preserved. Second, the structure retained for a local process is context dependent. 

This motivates using causal-beam conditions to constrain which task-level dependencies are preserved in the transferred representation, with contextual selection performed under the robot-side reference mechanisms rather than by directly reusing a human-side beam.

\begin{definition}[Causal Subtask Representation]
\label{def:causal_subtask}
Given a candidate manipulation SCM \(M\) for a local task process and a context \(u\), let \(M_u\) denote a causal beam of \(M\). We define the causal subtask representation as  
\begin{equation}
\sigma=(Y(u),\mathcal C_u,\mathcal R_u),
\end{equation}
where \(Y(u)\) denotes the realized local effect, \(\mathcal C_u=\{(C_i,C_i(u))\}_i\) collects the retained parents and their actual values, and \(\mathcal R_u\) records the parent–child relations retained within the projected local mechanisms. The projected mechanisms \(f_i^u\) are used to determine whether a retained parent set satisfies the causal-beam conditions, but the representation transfers the retained task-level dependencies rather than requiring identical projected functions across embodiments. Excluding a parent from the transferred dependency set does not remove its role in the projection context: its fixed value can still determine how the projected mechanism responds to changes in the retained inputs.
\end{definition}
The local task goal specifies the desired change, whereas \(\sigma\) describes dependencies supporting the observed local effect in a given context. An unrealized goal alone does not distinguish ongoing execution from failure.

\begin{figure}[t]
    \centering
    \includegraphics[width=1\linewidth]{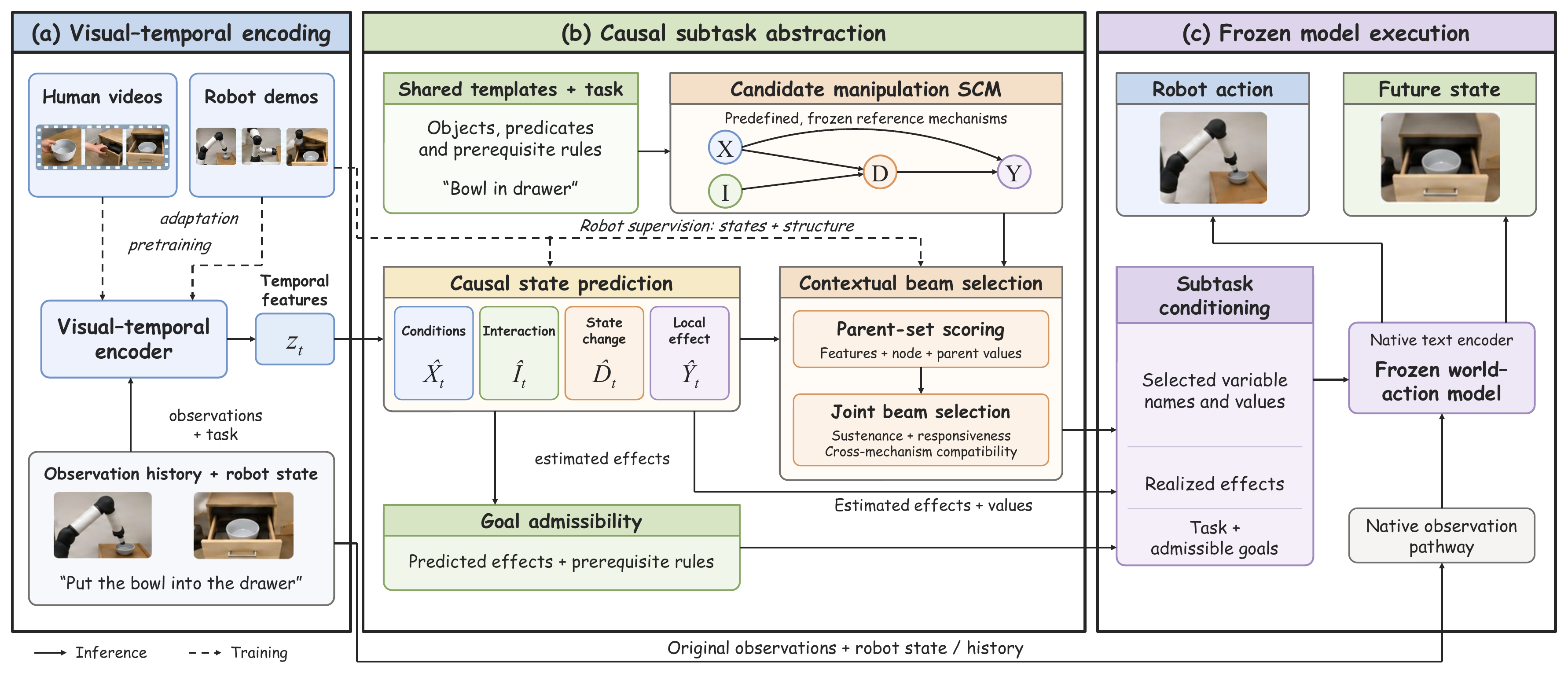}
    \caption{Overview of Causal-Beam-Guided Subtask Transfer (CBST). Human-video pretraining and robot adaptation support state prediction and contextual beam selection. The resulting semantic conditions guide a frozen world--action model to predict actions and future states. Solid and dashed arrows denote inference and training/supervision, respectively.}
    \label{fig:framework}
\end{figure}

\section{Method}
We propose Causal-Beam-Guided Subtask Transfer (CBST). Human-video pretraining initializes a visual–temporal module, which is then adapted with robot demonstrations to estimate task-level semantic variables from robot observations. Using these estimates, a learned selector chooses contextual parent assignments subject to causal-beam constraints within a predefined manipulation model. The selected assignments are rendered as language conditions for WAM, which combines them with the robot’s native observations to generate actions and future observation predictions.

\subsection{Overall Architecture}
Following the task-level formulation in Section~\ref{sec:causal_transfer_analysis}, CBST maps robot observations to semantic estimates, selects contextual dependencies, and uses the resulting subtask information to condition action generation. Given the interaction history \(h_t\) and task instruction \(g\), a visual--temporal encoder and semantic prediction heads produce
\begin{equation}
    z_t=E_\eta(h_t,g), \hat{\mathbf v}_t=P_\omega(z_t),
\end{equation}
where \(z_t\) summarizes the observed manipulation process and \(\hat{\mathbf v}_t\) estimates the conditions, manipulation attempts, object-state changes, and effects defined by the local SCM. Discretizing these predictions gives the contextual assignment \(\bar{\mathbf v}_t\). Predicted states and task-specified prerequisite rules also determine the currently admissible goals \(\hat G_t\).

The estimated context is then used to instantiate the causal-beam conditions in Section~\ref{sec:causal_beams}. Under fixed reference mechanisms \(f_i^{\mathrm{ref}}\), let \(\hat{\mathcal A}_{i,t}\) contain the parent subsets satisfying sustenance and responsiveness at \(\bar{\mathbf v}_t\). A learned selector assigns probabilities \(p_{i,t}(C)\) from \(z_t\) and the semantic predictions, and jointly selects compatible subsets:
\begin{equation}
\begin{aligned}
\{C_{i,t}^{*}\}_i\in \operatorname*{arg\,max}_{\{C_i\}_i} &\sum_i\log p_{i,t}(C_i)\\
\text{subject to}\quad &C_i\in\hat{\mathcal A}_{i,t}, \quad(\mathrm{PA}_i\setminus C_i)\cap C_j=\varnothing
\quad\text{for all }i\ne j.
\end{aligned}
\end{equation}
For a feasible selection, the retained assignments and relations form \(\hat{\sigma}_t=(\bar Y_t,\hat{\mathcal C}_t, \hat{\mathcal R}_t)\), an estimated instance of the causal subtask representation in Definition~\ref{def:causal_subtask}. Its validity is relative to the reference mechanisms and estimated context. A deterministic renderer combines the task instruction, admissible goals, and selected subtask information into a contextual description. The text encoder and conditional generative network complete the data flow:
\begin{equation}
    r_t=\mathcal R(g,\hat G_t,\hat{\sigma}_t), c_t=T_\theta(r_t), (A_t,\hat S_t^+)\sim p_\theta(\,\cdot\mid\mathcal O_t^W,c_t).
\end{equation}
Here, \(\mathcal O_t^W\) denotes the robot-observation condition, \(A_t\) the generated action chunk, and \(\hat S_t^+\) the predicted future observations. The renderer includes realized effects and selected variable assignments; retained relations guide field selection but are not explicitly serialized. If beam selection is infeasible, its fields are omitted while task and state conditioning are retained. The WAM and its text encoder remain frozen. After action execution, new observations update \(h_t\), and the estimation--selection--generation process repeats.

\subsection{Learning Objectives and Training}
Following the analysis in Section~\ref{sec:causal_transfer_analysis}, training grounds the task-level semantic variables and learns contextual parent selection under causal-beam constraints. We first initialize the visual--temporal encoder through action-category classification on human videos:
\begin{equation}
\mathcal L_h=\mathbb E_{(o^h,y^h)}\left[\operatorname{CE}\!\left(H(E_\eta(o^h)),y^h\right)\right],
\end{equation}
where \(o^h\) is a video clip, \(y^h\) its category label, and \(H\) a classification head. The pretrained encoder then initializes robot adaptation. Robot demonstrations supervise task states and interval variables through \(\mathcal L_{\mathrm{sem}}=\mathcal L_{\mathrm{state}}+\lambda_v\mathcal L_{\mathrm{var}}\), with each component normalized over valid labels. An auxiliary mechanism-consistency loss $\mathcal L_{\mathrm{mech}}$ encourages predicted parent variables to reproduce annotated outputs under the fixed reference mechanisms. This loss uses valid, reference-consistent annotations, while other valid labels remain available for semantic supervision. The reference mechanisms remain frozen during training.

Structural supervision follows the sustenance and responsiveness conditions in Section~\ref{sec:causal_beams}. For an annotated context \(v\), finite-domain enumeration under the fixed reference mechanisms produces the acceptable parent subsets \(\mathcal A_i(v)\). The selector learns to assign probability mass to this set, while a separate regularizer favors compact selections:
\begin{equation}
\mathcal L_{\mathrm{beam}}=-\sum_{i\in\mathcal I(v)}\log\sum_{C\in\mathcal A_i(v)}p_i(C),
\mathcal L_{\mathrm{size}}=\sum_{i\in\mathcal I(v)}\mathbb E_{C\sim p_i(\cdot\mid C\in\mathcal A_i(v))}[\frac{|C|}{|\mathrm{PA}_i|}].
\end{equation}
Here, \(\mathcal I(v)\) contains mechanisms with valid, reference-consistent annotations, complete variable domains, and nonempty acceptable sets. Other valid annotations remain available for semantic supervision. The beam loss permits multiple acceptable subsets; the size regularizer expresses a compactness preference rather than an additional beam-validity condition. Cross-mechanism compatibility is enforced by the joint selection procedure in the preceding subsection.
The robot adaptation objective is
\begin{equation}
\mathcal L_{\mathrm{robot}} = \mathcal L_{\mathrm{sem}}+\lambda_m\mathcal L_{\mathrm{mech}}+\lambda_b\mathcal L_{\mathrm{beam}} +\lambda_{\mathrm{size}}\mathcal L_{\mathrm{size}}.
\end{equation}
Training updates the trainable encoder components, semantic prediction heads, and selector. The reference mechanisms, WAM, and native text encoder remain frozen. At inference, beam validity is checked using the discrete estimated context, and compatible selections are used to construct the action-generation condition.

\begin{table}[t]
\centering

\begin{minipage}[t]{0.48\linewidth}
\centering
\renewcommand{\arraystretch}{1.1}
\caption{Success rates (\%) on RoboTwin 2.0..}
\label{tab:robotwin_main}
\begin{tabular}{@{}lcc@{}}
\toprule
Method & Easy & Hard \\
\midrule
X-VLA & 72.9 & 72.8 \\
$\pi_0$ & 65.9 & 58.4 \\
$\pi_{0.5}$ & 82.7 & 76.8 \\
Motus & 88.7 & 87.0 \\
\midrule
LingBot-VA & 92.93 & 91.55 \\
\textbf{LingBot-VA + CBST} & 94.00 & 92.54 \\
\bottomrule
\end{tabular}
\end{minipage}
\hfill
\begin{minipage}[t]{0.48\linewidth}
\centering
\caption{Success rates (\%) on LIBERO-Long.}
\label{tab:libero_main}
\begin{tabular}{@{}lc@{}}
\toprule
Method & Long SR (\%) \\
\midrule
X-VLA & 97.6 \\
Cosmos Policy (Predict2) & 97.6 \\
\midrule
LingBot-VA & $98.5\pm0.5$ \\
\textbf{LingBot-VA + CBST} & 98.60 \\
\midrule
Cosmos Policy (Predict2.5) & 96.6 \\
\textbf{Predict2.5 + CBST} & 97.33 \\
\bottomrule
\end{tabular}
\end{minipage}

\end{table}

\section{Experiments}\label{sec:experiments}

\subsection{Experimental Setup}
\label{sec:experimental_setup}
We pretrain CBST's ResNet18--GRU visual-temporal encoder on Something-Something V2 (SSv2)  \citep{ssv2dataset} and adapt the planner using robot demonstrations, keeping the WAM and its text encoder frozen. We evaluate LingBot-VA on RoboTwin 2.0 and LIBERO, and Cosmos Policy (Predict2.5) on LIBERO and RoboCasa \citep{robotwin2025,libero2023,robocasa2024}. Comparisons include the underlying WAMs and representative published policies, using officially reported baseline scores \citep{lingbotva2026,cosmoscookbook}. We report task success rates (\%) under official success criteria, averaged across evaluated tasks. Dataset composition, baseline descriptions, implementation and hardware configurations, and evaluation protocols are provided in Appendix~\ref{app:experimental_details}.

\subsection{Results on Simulation Benchmarks}
\label{sec:simulation_results}
\begin{figure}[t]
    \centering
    \includegraphics[width=\linewidth]{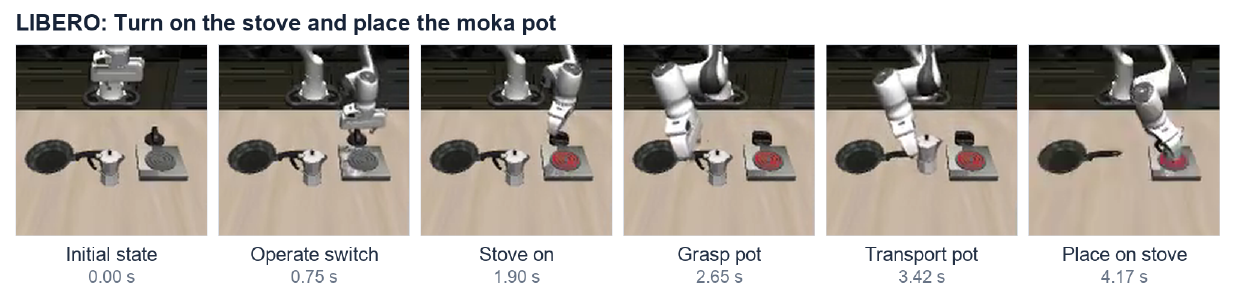}
    \caption{A LingBot-VA + CBST execution: operating the switch, turning on the stove, and placing the moka pot on it. Frames follow temporal order; timestamps indicate video playback time.}
    \label{fig:simulation_stove}
\end{figure}

We evaluate CBST with two frozen WAM backbones across three simulation benchmarks (Tables~\ref{tab:robotwin_main}--\ref{tab:robocasa_main}). On RoboTwin 2.0, LingBot-VA + CBST achieves 94.00\% and 92.54\% success under Easy and Hard settings. On LIBERO-Long, CBST achieves 98.60\% with LingBot-VA and 97.33\% with Cosmos Policy (Predict2.5). On RoboCasa, Predict2.5 + CBST reaches 72.44\% with 50 demonstrations per task, compared with 71.1\% for the backbone. Figure~\ref{fig:simulation_stove} illustrates a CBST execution on LIBERO. Full baseline comparisons, detailed results, and additional visualizations appear in Appendices~\ref{app:simulation_results}.

\begin{table}[t]
\centering

\begin{minipage}[t]{0.48\linewidth}
\centering
\renewcommand{\arraystretch}{1.2}
\caption{Success rates (\%) on RoboCasa.}
\label{tab:robocasa_main}
\resizebox{\linewidth}{!}{%
\begin{tabular}{@{}lcc@{}}
\toprule
Method & Demos/task & Average SR \\
\midrule
$\pi_0$ & 300 & 62.5 \\
GR00T-N1.5 & 300 & 64.1 \\
Video Policy & 300 & 66.0 \\
FLARE & 300 & 66.4 \\
Cosmos2 & 50 & 67.1 \\
\midrule
Cosmos2.5 & 50 & 71.1 \\
\textbf{Cosmos2.5 + CBST} & 50 & 72.44 \\
\bottomrule
\end{tabular}%
}
\end{minipage}
\hfill
\begin{minipage}[t]{0.48\linewidth}
\centering
\caption{Ablations on RoboTwin Easy.}
\label{tab:ablation_main}
\resizebox{\linewidth}{!}{%
\begin{tabular}{@{}llc@{}}
\toprule
Component & Variant & SR (\%) \\
\midrule
Reference & LingBot-VA & 92.93 \\
& \textbf{CBST} & 94.00 \\
\midrule
Pretraining & w/o human pretraining & 92.80 \\
& Shuffled frames & 92.94 \\
Conditioning & State-only & 92.82 \\
& Full structure & 93.42 \\
& Random selection & 92.56 \\
Objective & w/o $\mathcal L_{\mathrm{mech}}$ & 93.18 \\
\bottomrule
\end{tabular}%
}
\end{minipage}

\end{table}

\subsection{Ablation Studies}
\label{sec:ablations}
We examine human-video pretraining, contextual structural selection, and mechanism-consistency supervision on RoboTwin Easy with LingBot-VA as the frozen backbone (Table~\ref{tab:ablation_main}). Pretraining variants remove human-video initialization or shuffle its temporal order; conditioning variants use state-only information, the full candidate structure, or size-matched random subsets; the objective variant removes mechanism-consistency supervision. Robot adaptation data and evaluation settings remain fixed, while conditioning variants share the same trained predictor. More results across both datasets, detailed variant definitions, and computational costs are provided in Appendix~\ref{app:ablations}.

\subsection{Real-World Experiments}
\label{sec:real_world}
We evaluate CBST on an AgileX dual-arm platform across ten object-placement tasks, with 20 trials per task and method. Each task specifies an object and a fixed target location. In cross-workspace configurations, no single arm can reach both the object's initial position and its destination, requiring coordinated transport through a shared handover region. We compare Cosmos Policy with the Predict2.5 backbone against the same backbone augmented with CBST, and measure full-task success after object release. Task definitions, deployment settings, and the complete result table are provided in Appendix~\ref{app:realworld_details}. Figure~\ref{fig:realworld_cup} shows a recorded CBST cup-placement execution, including the intermediate handover. Additional sequences appear in Appendix~\ref{app:realworld_visualizations}.

\begin{figure}[t]
\centering
\includegraphics[width=\linewidth]{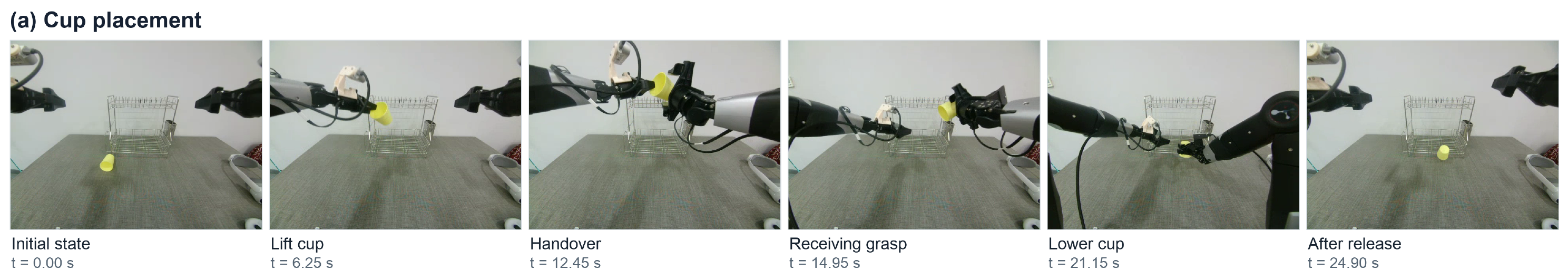}
\caption{CBST execution of cup placement on the AgileX platform. One arm lifts the cup and transfers it to the receiving gripper, which transports it to the target location and releases it on the tabletop. Handover enables cross-workspace transport but is not the task goal. Labels describe observed events, and timestamps indicate video playback time.}
\label{fig:realworld_cup}
\end{figure}

\subsection{Human-Video Pretraining}
\label{sec:human_pretraining}
\begin{figure}
    \centering
    \includegraphics[width=0.8\linewidth]{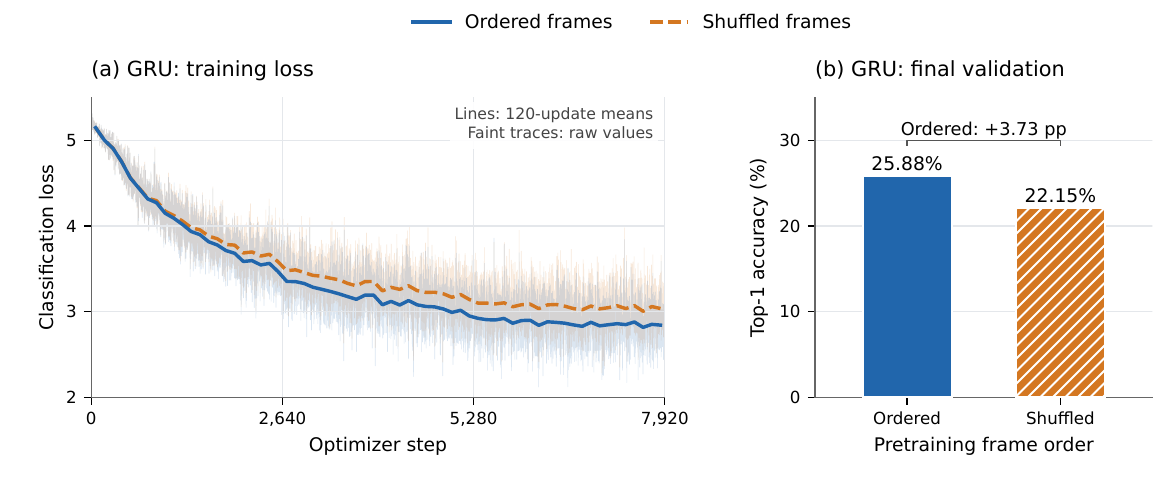}
    \caption{Temporal-order comparison for ResNet18--GRU pretraining on SSv2. (a) Training classification loss: bold lines show sample-weighted means over non-overlapping blocks of 120 updates; faint traces show the original values. (b) Final top-1 accuracy on 24,777 validation clips after three epochs, with one run per condition.}
\label{fig:human_pretraining}
\end{figure}

We examine the role of temporal ordering during human-video pretraining by comparing ordered and frame-shuffled SSv2 clips. Both variants use the same ResNet18--GRU architecture, action labels, sampled frame content, and optimization budget; only the temporal order of the pretraining inputs differs. Figure~\ref{fig:human_pretraining} shows the training classification loss and final validation top-1 accuracy after three epochs. Ordered pretraining achieves 25.88\% accuracy, compared with 22.15\% for the shuffled control, supporting the utility of temporal order for SSv2 action recognition.

We further evaluate the downstream contribution of this pretraining through the robot ablations in Table~\ref{tab:ablation_main}. Under the same robot adaptation protocol, ordered pretraining yields 94.00\% success on RoboTwin Easy, compared with 92.94\% for shuffled pretraining and 92.80\% without human-video pretraining. These comparisons support the practical benefit of ordered human-video pretraining for robot control. The pretrained representation initializes the robot planner; robot-specific prediction heads and structural selection are learned during adaptation.

\section{Conclusion}
We presented Causal-Beam-Guided Subtask Transfer (CBST), which uses a causal-beam perspective to guide contextual information selection for human-to-robot transfer. Human-video pretraining initializes manipulation-related representations, while robot demonstrations ground these representations in task-level semantic variables. Within a predefined manipulation model, causal-beam conditions constrain which contextual assignments are retained and supplied to a frozen world-action model for closed-loop control. Experiments on robot manipulation benchmarks show task-success improvements over the corresponding backbone baselines. 

\bibliography{main}
\bibliographystyle{plainnat}

\newpage

\appendix

\section*{AI Usage Statement}
In this work, we used generative AI tools for grammar and language polishing. They were not involved in research ideation, methodology, analysis, or substantive writing. The authors take full responsibility for all contents of the paper.

\section{Details of Causal Beams}\label{app:causal}

\subsection{Formal Details of Causal Beam Construction}

We consider a recursive structural causal model \(M=(U,V,\mathcal F)\), where \(U\) denotes the exogenous variables, \(V\) the endogenous variables, and \(\mathcal F\) the set of structural mechanisms. Each non-root endogenous variable \(V_i\) is determined by \(V_i=f_i(\mathrm{PA}_i,U)\), where \(\mathrm{PA}_i\subseteq V\) denotes the endogenous parents of \(V_i\). A context \(U=u\) determines the actual value \(V_i(u)\) of every endogenous variable. For a variable set \(Z\), we write \(Z_{W\leftarrow w}(u)\) for its value in the original model under the intervention \(W\leftarrow w\) and context \(u\).

\begin{definition}[Causal Beam {\citep{pearl2009causality}}]
\label{def:causal_beam_appendix}
Given a recursive structural causal model \(M=(U,V,\mathcal F)\) and a context \(u\), a causal beam \(M_u=(u,V,\mathcal F^u)\) is constructed by projecting the structural mechanisms as follows.

For each projected non-root mechanism \(V_i=f_i(\mathrm{PA}_i,U)\), partition its parent set as \(\mathrm{PA}_i=C_i\mathbin{\dot\cup}N_i\), where \(C_i\) denotes the retained parents and \(N_i\) the remaining parents. The actual values of the retained parents must sustain the actual output:
\begin{equation}
f_i(C_i(u),n_i,u)=V_i(u),
\qquad
\forall n_i\in\operatorname{Val}(N_i).
\end{equation}

Next, select \(W_i\subseteq N_i\) and an assignment \(w_i\). Let \(N_{i,W_i\leftarrow w_i}(u)\) denote the values attained by the remaining parents \(N_i\) in the original model under the intervention \(W_i\leftarrow w_i\) and context \(u\). The projected mechanism is
\begin{equation}
f_i^u(c_i)=f_i\!\left(c_i,N_{i,W_i\leftarrow w_i}(u),u\right).
\end{equation}

The projected mechanism must remain nontrivially responsive to the retained inputs: there exists \(c_i'\in\operatorname{Val}(C_i)\) such that \(f_i^u(c_i')\neq V_i(u)\). Thus, the projection cannot simply replace the original mechanism with a constant function that always returns the actual output.

Selections across different mechanisms must additionally be compatible. In particular, \(N_i\cap C_j=\varnothing\) for \(i\neq j\), and whenever intervention assignments overlap across mechanisms, they must assign the same value to every shared intervened variable. The compatible projected mechanisms, together with the retained assignment rules for root variables, constitute the causal beam \(M_u=(u,V,\mathcal F^u)\).
\end{definition}

The sustaining condition requires the actual values \(C_i(u)\) to preserve the actual mechanism output under arbitrary assignments to the remaining parents \(N_i\). The responsiveness condition serves a different role: it requires the projected mechanism to preserve a nontrivial dependence on its retained inputs. These two requirements distinguish a causal projection from a representation that merely records the observed output.

A special case is obtained when no non-actual contingency is introduced.

\begin{definition}[Natural Beam {\citep{pearl2009causality}}]
\label{def:natural_beam_appendix}
A causal beam is called \emph{natural} when \(W_i=\varnothing\) for every projected mechanism. In this case, the remaining parents are fixed at their actual values under context \(u\), and the projected mechanism reduces to \(f_i^u(c_i)=f_i(c_i,N_i(u),u)\).
\end{definition}

A variable that is fixed in a particular projection is not thereby declared irrelevant in general. The projection is defined relative to the specified model and context, and a different context may admit a different set of retained parents and hence a different projected mechanism.

\subsection{An Illustrative Construction of a Causal Beam}

We illustrate the construction using a binary structural causal model with three root variables \(B,G,O\) and two non-root variables \(D,Y\), with mechanisms \(D=B\land G\) and \(Y=D\land O\). The parent sets are therefore \(\mathrm{PA}_D=\{B,G\}\) and \(\mathrm{PA}_Y=\{D,O\}\).

We first consider a context \(u_1\) with \(B=1,G=1,O=1\), which implies \(D=1,Y=1\). For the mechanism \(D=B\land G\), retaining only \(B\) does not sustain the actual output because setting \(G=0\) changes \(D\) from \(1\) to \(0\). Likewise, retaining only \(G\) fails because setting \(B=0\) changes the output. A valid choice is therefore \(C_D=\{B,G\}\) and \(N_D=\varnothing\). The corresponding projection is \(f_D^{u_1}(b,g)=b\land g\). It preserves the actual output at \(b=g=1\) and remains nontrivially responsive because, for example, \(f_D^{u_1}(0,1)=0\neq1=D(u_1)\).

For \(Y=D\land O\), the same reasoning gives \(C_Y=\{D,O\}\) and \(N_Y=\varnothing\), with \(f_Y^{u_1}(d,o)=d\land o\). Again, the actual output is preserved, while changing either retained input appropriately can change the projected output. Because both excluded parent sets are empty, the cross-mechanism compatibility condition is satisfied. The resulting natural beam therefore retains the original mechanisms for this context.

Now consider a second context \(u_2\) with \(B=1,G=0,O=1\), which gives \(D=0,Y=0\). For \(D=B\land G\), choose \(C_D=\{G\}\) and \(N_D=\{B\}\). The sustaining condition holds because \(f_D(b,0)=b\land0=0\) for all \(b\in\{0,1\}\). Using a natural projection fixes \(B\) at its actual value \(B(u_2)=1\), yielding \(f_D^{u_2}(g)=f_D(1,g)=g\). The projection remains nontrivially responsive since \(f_D^{u_2}(1)=1\neq0=D(u_2)\).

For \(Y=D\land O\), choose \(C_Y=\{D\}\) and \(N_Y=\{O\}\). The sustaining condition is satisfied because \(f_Y(0,o)=0\land o=0\) for all \(o\in\{0,1\}\). Fixing \(O\) at its actual value \(O(u_2)=1\) gives \(f_Y^{u_2}(d)=f_Y(d,1)=d\), which is also nontrivially responsive because \(f_Y^{u_2}(1)=1\neq0=Y(u_2)\).

The cross-mechanism compatibility conditions are satisfied because \(N_D\cap C_Y=\{B\}\cap\{D\}=\varnothing\) and \(N_Y\cap C_D=\{O\}\cap\{G\}=\varnothing\). No non-actual intervention assignments are introduced, so there are no additional assignment-consistency constraints to check. The resulting natural beam contains the projected mechanisms \(D=G\) and \(Y=D\).

The two contexts therefore yield different valid projections of the same underlying structural causal model. In \(u_1\), sustaining the actual successful outputs requires retaining both parents of each mechanism, whereas in \(u_2\), the actual zero values allow smaller retained parent sets. The construction is context specific: fixing \(B\) and \(O\) in the second projection does not imply that these variables are generally irrelevant, only that their actual values can be treated as fixed within that particular projection.

\section{Experimental Details}
\label{app:experimental_details}

\subsection{Datasets and Benchmark Composition}
\label{app:datasets}

We distinguish the data used to pretrain the planner from the robot demonstrations used for adaptation and the environments used for closed-loop evaluation. Table~\ref{tab:dataset_overview} summarizes their roles.

\begin{table}[t]
\centering

\caption{Datasets and evaluation scope. The RoboCasa entry refers to the 24-task protocol used by Cosmos Policy, rather than the complete RoboCasa task collection.}
\label{tab:dataset_overview}
\begin{tabularx}{\linewidth}{@{}lXX@{}}
\toprule
Dataset & Role in CBST & Scope / WAM \\
\midrule
SSv2 & Human-video representation pretraining & 174 action categories; planner initialization \\
RoboTwin 2.0 & Robot adaptation and evaluation & 50 tasks; LingBot-VA \\
LIBERO-Long & Robot adaptation and evaluation & 10 tasks; LingBot-VA and Cosmos Policy (Predict2.5) \\
RoboCasa & Robot adaptation and evaluation & 24 tasks; Cosmos Policy (Predict2.5) \\
\bottomrule
\end{tabularx}
\end{table}

\subsubsection{Something-Something V2}
\label{app:ssv2}

SSv2 contains short videos of everyday human--object interactions \citep{ssv2dataset}. Its 174 labels describe actions and their outcomes using object-independent templates. The official release contains 220,847 videos, divided into 168,913 training videos, 24,777 validation videos, and 27,157 test videos. Test labels are not publicly provided.

\begin{table}[t]
\centering
\caption{Official SSv2 split sizes. These are release-level counts; the number of clips retained after preprocessing is reported separately in Table~\ref{tab:training_configuration}.}
\label{tab:ssv2_splits}
\begin{tabular}{@{}lrl@{}}
\toprule
Split & Videos & Annotation availability \\
\midrule
Training & 168,913 & Action labels \\
Validation & 24,777 & Action labels \\
Test & 27,157 & Labels withheld \\
\midrule
Total & 220,847 & 174 action categories \\
\bottomrule
\end{tabular}
\end{table}

\paragraph{Action coverage.}
The label space includes object displacement, placement, containment, deformation, and interactions with different observable outcomes. Table~\ref{tab:ssv2_action_overview} summarizes representative distinctions relevant to manipulation. The groups are descriptive summaries, not an alternative official taxonomy. Training retains the original 174-class classification target.

\begin{table}[t]
\centering
\caption{Representative action distinctions in SSv2. Descriptions are abbreviated semantic summaries rather than verbatim labels or an exhaustive class list.}
\label{tab:ssv2_action_overview}
\begin{tabularx}{\linewidth}{@{}p{0.25\linewidth}X@{}}
\toprule
Interaction type & Representative distinctions \\
\midrule
Object displacement & Directional movement; motion toward or away from the camera. \\
Placement and removal & Moving objects onto surfaces, into receptacles, or away from their previous locations. \\
Lifting and release & Partial versus complete lifting; holding versus releasing. \\
Contact and motion & Contact that moves an object versus contact with little observable displacement. \\
Support and stability & Tilting a support with or without causing the supported object to fall. \\
Attachment & Successful attachment versus an unsuccessful attempt. \\
Deformation & Bending, breaking, folding, unfolding, and squeezing. \\
Visibility & Covering or uncovering an object. \\
Object interaction & Collisions with different resulting motions. \\
Action completion & Actual execution versus pretending or attempting without corresponding effect. \\
\bottomrule
\end{tabularx}
\end{table}

\paragraph{Use in CBST.}
Human pretraining optimizes action classification. The visual encoder, temporal module, and representation projection initialize the robot-side planner; robot prediction heads and structural selectors are introduced during adaptation. Human category labels are not treated as robot action labels or as direct supervision for causal parent sets.

For the temporal-order control, sampled frames are shuffled within each clip while preserving the clip label, sampled frame content, and training budget. The no-human-pretraining variant uses the same robot adaptation procedure without SSv2 pretraining. Its visual initialization is specified separately to distinguish absence of human-video pretraining from random initialization.

\subsubsection{RoboTwin 2.0}
\label{app:robotwin}

RoboTwin 2.0 provides simulated bimanual manipulation tasks \citep{robotwin2025}. We use the 50-task evaluation set reported for LingBot-VA. The tasks cover object transport, handover, placement, stacking, tool interaction, and articulated-object manipulation. Easy and Hard results are reported separately.

Table~\ref{tab:robotwin_task_groups} lists the complete task set, grouped by the horizon annotations in the LingBot-VA evaluation table \citep{lingbotva2026}. These annotations describe task groups; they are distinct from the action-chunk length and the simulator episode limit.

\begin{table}[t]
\centering

\caption{The 50 RoboTwin tasks grouped by the horizon labels used in the LingBot-VA report. Names are abbreviated from the official task identifiers.}
\label{tab:robotwin_task_groups}
\begin{tabularx}{\linewidth}{@{}ccX@{}}
\toprule
Horizon & Tasks & Task names \\
\midrule
1 & 30 & Adjust Bottle; Beat Block Hammer; Click Alarmclock; Click Bell; Dump Bin Bigbin; Grab Roller; Lift Pot; Move Can Pot; Move Pillbottle Pad; Move Playingcard Away; Move Stapler Pad; Open Laptop; Open Microwave; Place A2B Left; Place A2B Right; Place Bread Basket; Place Container Plate; Place Empty Cup; Place Fan; Place Mouse Pad; Place Object Scale; Place Object Stand; Place Phone Stand; Place Shoe; Press Stapler; Rotate QRcode; Shake Bottle Horizontally; Shake Bottle; Stamp Seal; Turn Switch. \\
\addlinespace
2 & 15 & Handover Block; Handover Mic; Hanging Mug; Pick Diverse Bottles; Pick Dual Bottles; Place Bread Skillet; Place Burger Fries; Place Can Basket; Place Cans Plasticbox; Place Dual Shoes; Place Object Basket; Put Object Cabinet; Scan Object; Stack Blocks Two; Stack Bowls Two. \\
\addlinespace
3 & 5 & Blocks Ranking RGB; Blocks Ranking Size; Put Bottles Dustbin; Stack Blocks Three; Stack Bowls Three. \\
\bottomrule
\end{tabularx}
\end{table}

\paragraph{Evaluation settings.}
Easy and Hard follow the configurations associated with the reported LingBot-VA results. Hard introduces additional variation in object poses and scene configurations. We retain the benchmark-defined success checks. Overall performance averages the task success rates over all 50 tasks; horizon-specific results use only the tasks in the corresponding group.

The robot embodiment, camera configuration, demonstration budget, and episode limits used for CBST are specified in Tables~\ref{tab:training_configuration} and~\ref{tab:evaluation_configuration}.

\subsubsection{LIBERO}
\label{app:libero}

LIBERO evaluates knowledge transfer in language-conditioned robot manipulation \citep{libero2023}. Its released collections include Spatial, Object, Goal, and LIBERO-100. LIBERO-100 is partitioned into LIBERO-90 and the 10-task LIBERO-Long set, also registered as \texttt{libero\_10}.

We evaluate CBST on all four LIBERO suites—Spatial, Object, Goal, and Long—with both WAM backbones. Each suite contains ten tasks. The main text focuses on LIBERO-Long, while Appendix C.2 reports results on all four suites. Table~\ref{tab:libero_suite_overview} distinguishes these evaluation suites from the broader data collection.

\begin{table}[t]
\centering

\caption{LIBERO suite composition. LIBERO-90 and LIBERO-Long are the two partitions of LIBERO-100, not additional copies of that collection.}
\label{tab:libero_suite_overview}
\begin{tabularx}{\linewidth}{@{}lcX@{}}
\toprule
Suite & Tasks & Main variation \\
\midrule
Spatial & 10 & Spatial relations identifying a target object. \\
Object & 10 & Object identity under a shared placement goal. \\
Goal & 10 & Desired outcomes within related environments. \\
LIBERO-90 & 90 & Diverse manipulation tasks. \\
Long / LIBERO-10 & 10 & Tasks requiring extended execution. \\
\bottomrule
\end{tabularx}
\end{table}

\paragraph{Spatial, Object, and Goal.}
Table~\ref{tab:libero_short_suite_tasks} summarizes the task variants in the three short-horizon suites. LIBERO provides scene and language specifications together with visual observations and robot proprioception. The released demonstrations include workspace and wrist views.

\begin{table}[t]
\centering

\caption{Task composition of the Spatial, Object, and Goal suites. Shared instructions are factored out for compactness.}
\label{tab:libero_short_suite_tasks}
\begin{tabularx}{\linewidth}{@{}p{0.17\linewidth}X@{}}
\toprule
Suite & Task variants \\
\midrule
Spatial & Move the black bowl to the plate. The bowl starts: between plate and ramekin; beside ramekin; at table center; on cookie box; inside top drawer; on ramekin; beside cookie box; on stove; beside plate; on cabinet. \\
\addlinespace
Object & Place each target in the basket: alphabet soup; cream cheese; salad dressing; barbecue sauce; ketchup; tomato sauce; butter; milk; chocolate pudding; orange juice. \\
\addlinespace
Goal & Open middle drawer; bowl to stove; wine bottle to cabinet top; bowl inside top drawer; bowl to cabinet top; plate toward stove front; cream cheese into bowl; activate stove; bowl to plate; wine bottle to rack. \\
\bottomrule
\end{tabularx}
\end{table}

\paragraph{Long-horizon tasks.}
Table~\ref{tab:libero_long_tasks} lists all ten tasks used for CBST evaluation. Task identifiers and success predicates follow the official benchmark; the descriptions below are shortened for readability.

\begin{table}[t]
\centering

\caption{The ten LIBERO-Long tasks. LR, K, and S denote living-room, kitchen, and study scenes, respectively.}
\label{tab:libero_long_tasks}
\begin{tabularx}{\linewidth}{@{}clX@{}}
\toprule
ID & Scene & Goal summary \\
\midrule
L1 & LR2 & Basket: alphabet soup and tomato sauce. \\
L2 & LR2 & Basket: cream cheese and butter. \\
L3 & K3 & Activate stove; place moka pot on it. \\
L4 & K4 & Black bowl inside bottom drawer; drawer closed. \\
L5 & LR5 & White mug on left plate; yellow-and-white mug on right plate. \\
L6 & S1 & Book in caddy's rear compartment. \\
L7 & LR6 & White mug on plate; pudding to its right. \\
L8 & LR1 & Basket: alphabet soup and cream cheese. \\
L9 & K8 & Both moka pots on stove. \\
L10 & K6 & Yellow-and-white mug inside microwave; door closed. \\
\bottomrule
\end{tabularx}
\end{table}

\paragraph{Data and evaluation.}
Robot demonstrations are partitioned by trajectory before observation windows are constructed. Planner inputs use observations available up to the current decision time, together with robot proprioception and task information. Simulator state may provide adaptation labels, but is not an input to the learned planner during evaluation.

We retain the official task success predicates. Success rates are averaged over the ten Long tasks. Scores for the complete four-suite benchmark, when quoted from prior work, retain the aggregation reported by that source.

\subsubsection{RoboCasa}
\label{app:robocasa}

RoboCasa provides kitchen environments for learning everyday manipulation skills \citep{robocasa2024}. We use the 24-task protocol adopted by Cosmos Policy \citep{cosmospolicy2026}. This evaluation set contains seven task families, listed in Table~\ref{tab:robocasa_tasks}; it excludes kitchen navigation.

\begin{table}[t]
\centering

\caption{The 24 RoboCasa tasks used in the Cosmos Policy evaluation protocol. Task identifiers follow its environment release. PnP denotes pick-and-place.}
\label{tab:robocasa_tasks}
\begin{tabularx}{\linewidth}{@{}p{0.17\linewidth}cX@{}}
\toprule
Family & Count & Tasks \\
\midrule
Pick-and-place & 8 & PnPCounterToCab; PnPCabToCounter; PnPCounterToSink; PnPSinkToCounter; PnPCounterToMicrowave; PnPMicrowaveToCounter; PnPCounterToStove; PnPStoveToCounter. \\
\addlinespace
Cabinet doors & 4 & OpenSingleDoor; CloseSingleDoor; OpenDoubleDoor; CloseDoubleDoor. \\
\addlinespace
Drawers & 2 & OpenDrawer; CloseDrawer. \\
\addlinespace
Sink & 3 & TurnOnSinkFaucet; TurnOffSinkFaucet; TurnSinkSpout. \\
\addlinespace
Stove & 2 & TurnOnStove; TurnOffStove. \\
\addlinespace
Coffee & 3 & CoffeeSetupMug; CoffeeServeMug; CoffeePressButton. \\
\addlinespace
Microwave & 2 & TurnOnMicrowave; TurnOffMicrowave. \\
\bottomrule
\end{tabularx}
\end{table}

\paragraph{Task coverage.}
Pick-and-place tasks vary the source and destination fixture. Door and drawer tasks require changing articulated states. Sink, stove, and microwave tasks involve fixture controls, whereas coffee tasks cover mug positioning, retrieval, and button operation.

\paragraph{Reference protocol.}
The Cosmos Policy report uses 50 human-teleoperated demonstrations per task. Its evaluation includes five scenes per task, with ten trials per scene and three evaluation seeds, giving 3,600 trials over the 24 tasks. Evaluation uses unseen object instances, and two of the five scenes introduce styles absent from training.

These quantities describe the published reference protocol. CBST's adaptation data and rollout counts are specified separately below. Demonstration budgets remain explicit in the result tables because the published baselines use different amounts of data.

\subsection{Baseline Details}
\label{app:baselines}

We include two types of comparison methods: the WAM backbones to which CBST is attached, and external robot manipulation methods reported on the same benchmarks. Only LingBot-VA and Cosmos Policy (Predict2.5) serve as CBST backbones.

\paragraph{LingBot-VA.}
LingBot-VA jointly generates visual futures and robot actions using an autoregressive video--action model \citep{lingbotva2026}. We use it as the underlying WAM for RoboTwin and LIBERO-Long.

\paragraph{Cosmos Policy.}
Cosmos Policy adapts a pretrained video model to generate actions together with future-state and value predictions \citep{cosmospolicy2026}. CBST uses the Predict2.5 configuration. The earlier Predict2 configuration is retained as an additional published reference, with its results identified separately \citep{cosmoscookbook}.

\paragraph{X-VLA.}
X-VLA uses a flow-based action model and soft prompts to accommodate heterogeneous embodiments and data sources \citep{xvla2025}.

\paragraph{$\pi_0$ and $\pi_{0.5}$.}
$\pi_0$ combines pretrained vision--language representations with flow-based action generation \citep{pi02024}. $\pi_{0.5}$ extends this line with heterogeneous training sources and semantic supervision for broader generalization \citep{pi052025}.

\paragraph{Motus.}
Motus combines understanding, video generation, and action modeling within a unified latent-action framework \citep{motus2025}.

\paragraph{OpenVLA-OFT.}
OpenVLA-OFT applies an optimized adaptation recipe to OpenVLA, including parallel action decoding, action chunking, and continuous action prediction \citep{openvlaoft2025}.

\paragraph{GR00T-N1.5.}
GR00T-N1.5 combines vision--language encoding with a diffusion action module and updates the model and training recipe of GR00T-N1 \citep{gr00tn12025}.

\paragraph{Video Policy.}
Video Policy couples video generation with action generation, using visual prediction to support robot policy learning \citep{videopolicy2025}.

\begin{table}[t]
\centering

\caption{Roles of the principal comparison methods. R, L, and C denote RoboTwin, LIBERO, and RoboCasa. Benchmark entries indicate the comparisons included in this paper, not the complete capabilities of each method.}
\label{tab:baseline_overview}
\begin{tabularx}{\linewidth}{@{}lXlc@{}}
\toprule
Method & Model / interface & Role & Benchmark \\
\midrule
LingBot-VA & Video--action generation & Backbone & R, L \\
Cosmos Policy--Predict2.5 & Joint action and future prediction & Backbone & L, C \\
Cosmos Policy--Predict2 & Earlier video-model configuration & Reference & L, C \\
X-VLA & Soft-prompted VLA & Reference & R, L \\
$\pi_0$ & Flow-based VLA & Reference & R, L, C \\
$\pi_{0.5}$ & Generalist VLA & Reference & R, L \\
Motus & Unified latent-action world model & Reference & R \\
OpenVLA-OFT & Adapted VLA with action chunking & Reference & L \\
GR00T-N1.5 & Vision--language and diffusion action model & Reference & C \\
Video Policy & Coupled video and action generation & Reference & C \\
\bottomrule
\end{tabularx}
\end{table}

\paragraph{Sources of reported results.}
RoboTwin comparisons follow the LingBot-VA report. LIBERO comparisons combine the published LingBot-VA and Cosmos Policy tables. RoboCasa comparisons follow Cosmos Policy, with Predict2.5 results obtained from the official Cosmos Cookbook.

We preserve the reported model version, demonstration budget, evaluation setting, and uncertainty notation. When two sources report different results for the same method, the corresponding source is identified explicitly; entries from different sources are not combined into a single synthetic result.

\subsection{Implementation Details}
\label{app:implementation}

\paragraph{Visual-temporal encoder.}
The planner uses a ResNet18 visual encoder \citep{resnet2016} and a GRU temporal module \citep{gru2014}. Available camera views share the visual encoder, and their features are averaged at each sampled timestep. The resulting feature is combined with robot proprioception and task features before temporal aggregation. Observation histories combine sparse earlier observations with denser recent observations.

\paragraph{Human pretraining and robot adaptation.}
Human pretraining uses action-category classification. The pretrained encoder and representation projection initialize the robot planner, while robot-specific semantic heads and parent-set selectors are newly initialized.

During robot adaptation, the final visual stage, temporal module, representation projection, semantic heads, and selectors are optimized using the objectives defined in the method section. The earlier visual stages, reference mechanisms, WAM, and native text encoder remain frozen. Trajectory splits are constructed before sampling observation windows.

\paragraph{Semantic and structural supervision.}
Robot demonstration replay provides labels for semantic states, interval variables, and realized effects. Unknown labels are masked rather than assigned negative values. Candidate dependencies are instantiated from the task specification and shared manipulation templates.

For structural supervision, parent subsets are evaluated within the declared finite variable domains using frozen reference mechanisms. Multiple acceptable subsets are retained when they satisfy the structural conditions. Examples lacking the required labels or variable domains are excluded from the corresponding structural loss.

\paragraph{Conditioning and execution.}
A deterministic renderer preserves the original task instruction and appends predicted task information and selected dependency fields. The WAM's native text encoder processes the resulting prompt. The WAM retains its original observation encoder, action representation, and generation procedure.

The planner updates the condition from newly available observations before the next action request. Evaluation does not provide the learned planner with future observations or simulator ground-truth semantic states. Any oracle condition is identified separately.

\begin{table}[t]
\centering
\caption{Planner architecture and training configuration.}
\label{tab:training_configuration}
\begin{tabularx}{\linewidth}{@{}p{0.43\linewidth}X@{}}
\toprule
Item & Configuration \\
\midrule
Visual encoder
& ResNet18 \\
Temporal encoder
& GRU \\
Camera-feature fusion
& Mean pooling over available camera views \\
Human pretraining
& 174-class action classification on SSv2
using cross-entropy loss \\
Transferred components
& Visual encoder, temporal module, and
representation projection \\
Robot prediction heads
& Semantic states, interval variables,
and candidate parent subsets \\
Observation history
& Observed-prefix inputs with sparse earlier
observations and dense recent observations \\
Robot adaptation
& Update the final visual stage, temporal module,
representation projection, semantic heads,
and parent-set selectors \\
Frozen components
& Earlier visual stages, reference mechanisms,
WAM, and native text encoder \\
Conditioning interface
& Deterministic text rendering followed by
the WAM's native text encoder \\
Robot data split
& Split by demonstration trajectory before
constructing observation windows \\
Missing supervision
& Mask unknown labels in the corresponding losses \\
\bottomrule
\end{tabularx}
\end{table}



\subsection{Evaluation Metrics and Protocols}
\label{app:evaluation}

\paragraph{Task success.}
Let $s_{k,e}\in\{0,1\}$ denote the official success indicator for episode $e$ of task $k$, with $N_k$ evaluated episodes. The task success rate is
\begin{equation}
    \mathrm{SR}_k = \frac{100}{N_k}\sum_{e=1}^{N_k}s_{k,e}.
\end{equation}
The aggregate score over $K$ tasks is
\begin{equation}
    \mathrm{SR}_{\mathrm{avg}} = \frac{1}{K}\sum_{k=1}^{K}\mathrm{SR}_k.
\end{equation}
When every task has the same number of episodes, this equals the success rate pooled over all episodes.

RoboTwin Easy and Hard are aggregated separately. LIBERO-Long averages over ten tasks, and the RoboCasa protocol averages over 24 tasks. Published baseline values retain the aggregation and uncertainty conventions of their respective sources.

\paragraph{Evaluation configurations.}
Table~\ref{tab:evaluation_configuration} records the protocol for each CBST evaluation. Action-chunk length, executed actions per request, and episode limits are specified separately because they determine different aspects of closed-loop execution.

\begin{table}[t]
\centering
\caption{Evaluation scope and episode budgets. RoboTwin counts apply separately to Easy and Hard. LIBERO-Long and RoboCasa counts aggregate evaluation seeds. Each WAM is evaluated independently.}
\label{tab:evaluation_configuration}
\begin{tabularx}{\linewidth}{@{}p{0.25\linewidth}XXXX@{}}
\toprule
& RoboTwin 2.0 & \multicolumn{2}{c}{LIBERO-Long} & RoboCasa \\
\cmidrule(lr){3-4}
Item & LingBot-VA & LingBot-VA & Predict2.5 & Predict2.5 \\
\midrule
Tasks & 50 & 10 & 10 & 24 \\
Episodes per task & 100 & 150 & 150 & 150 \\
Total episodes & 5,000 & 1,500 & 1,500 & 3,600 \\
\bottomrule
\end{tabularx}
\end{table}

\paragraph{Training and evaluation randomness.}
Training seeds and evaluation seeds are recorded separately. Repeated rollout seeds assess variation in evaluation, whereas independently trained planners assess variation in learning. The meaning of any reported error bar is specified in the corresponding table or figure caption.

\section{Additional Simulation Results}
\label{app:simulation_results}

This section expands the benchmark comparisons in Section~\ref{sec:simulation_results}. Published values retain their original precision and aggregation.

\subsection{RoboTwin 2.0}
\label{app:robotwin_results}

Table~\ref{tab:robotwin_horizon} reports the horizon-grouped results \citep{lingbotva2026}. Easy and Hard results remain separate. Source-reported aggregate values are retained without recomputing them from the rounded per-task entries.

\begin{table}[t]
\centering
\caption{RoboTwin success rates (\%) grouped by task horizon. Each cell reports Easy / Hard. Values follow Table~1 of \citep{lingbotva2026}; X-VLA$^{*}$ is adopted there from Motus.}
\label{tab:robotwin_horizon}
\begin{tabular}{@{}lcccc@{}}
\toprule
Method & Horizon 1 & Horizon 2 & Horizon 3 & All 50 tasks \\
\midrule
X-VLA$^{*}$ & 81.6 / 82.5 & 59.3 / 55.9 & 61.2 / 66.0 & 72.9 / 72.8 \\
$\pi_0$ & 66.5 / 61.6 & 66.1 / 54.7 & 61.6 / 50.2 & 65.9 / 58.4 \\
$\pi_{0.5}$ & 85.1 / 80.2 & 79.3 / 73.0 & 78.6 / 67.4 & 82.7 / 76.8 \\
Motus & 91.0 / 90.6 & 85.2 / 80.9 & 85.0 / 84.2 & 88.7 / 87.0 \\
LingBot-VA & 94.18 / 93.56 & 90.35 / 86.95 & 93.22 / 93.28 & 92.93 / 91.55 \\
\midrule
\textbf{LingBot-VA + CBST} & 94.73 / 93.97 & 91.93 / 88.80 & 95.80 / 95.20 & 94.00 / 92.54 \\
\bottomrule
\end{tabular}
\end{table}

\subsection{LIBERO}
\label{app:libero_results}

Table~\ref{tab:libero_complete} combines the four-suite comparisons reported by LingBot-VA and Cosmos Policy, together with the official Predict2.5 results. Identical entries are consolidated, while source-dependent differences are preserved as separate rows. In particular, the reported $\pi_0$ averages and UniVLA results differ between the two papers.CBST results are reported for Spatial, Object, Goal, and Long with both WAM backbones. For CBST, the Average column is the arithmetic mean of the four suite success rates; source-reported baseline averages are preserved.

\begin{table}[t]
\centering
\caption{Complete published LIBERO comparison. All values are success rates (\%). L: Table~2 of \citep{lingbotva2026}; C: Table~1 of \citep{cosmospolicy2026}; K: official Cosmos Cookbook \citep{cosmoscookbook}. Source-reported averages and uncertainty values are preserved.}
\label{tab:libero_complete}
\resizebox{0.95\textwidth}{!}{
\begin{tabular}{@{}lcccccc@{}}
\toprule
Method & Source & Spatial & Object & Goal & Long & Average \\
\midrule
Octo & L & 78.9 & 85.7 & 84.6 & 51.1 & 75.1 \\
Seer & L & -- & -- & -- & 87.7 & -- \\
MoDE & L & -- & -- & -- & 94.0 & -- \\
SuSIE & L & -- & -- & -- & 76.3 & -- \\
SpatialVLA & L & 88.2 & 89.9 & 78.6 & 55.5 & 78.1 \\
TraceVLA & L & 84.6 & 85.2 & 75.1 & 54.1 & 74.8 \\
CoT-VLA & L & 87.5 & 91.6 & 87.6 & 69.0 & 81.1 \\
ThinkAct & L & 88.3 & 91.4 & 87.1 & 70.9 & 84.4 \\
SmolVLA & L & 93.0 & 94.0 & 91.0 & 77.0 & 88.8 \\
CronusVLA & L & 97.3 & 99.6 & 96.9 & 94.0 & 97.0 \\
FLOWER & L & 97.1 & 96.7 & 95.6 & 93.5 & 95.7 \\
GR00T-N1 & L & 94.4 & 97.6 & 93.0 & 90.6 & 93.9 \\
$\pi_0$ & L & 96.8 & 98.8 & 95.8 & 85.2 & 94.1 \\
$\pi_0$ & C & 96.8 & 98.8 & 95.8 & 85.2 & 94.2 \\
$\pi_0$+FAST & L & 96.4 & 96.8 & 88.6 & 60.2 & 85.5 \\
$\pi_{0.5}$ & C & 98.8 & 98.2 & 98.0 & 92.4 & 96.9 \\
OpenVLA & L & 84.7 & 88.4 & 79.2 & 53.7 & 76.5 \\
OpenVLA-OFT & L, C & 97.6 & 98.4 & 97.9 & 94.5 & 97.1 \\
DD-VLA & L & 97.2 & 98.6 & 97.4 & 92.0 & 96.3 \\
UniVLA & L & 95.4 & 98.8 & 93.6 & 94.0 & 95.4 \\
UniVLA & C & 96.5 & 96.8 & 95.6 & 92.0 & 95.2 \\
X-VLA & L & 98.2 & 98.6 & 97.8 & 97.6 & 98.1 \\
Diffusion Policy & C & 78.3 & 92.5 & 68.3 & 50.5 & 72.4 \\
Dita & C & 97.4 & 94.8 & 93.2 & 83.6 & 92.3 \\
UVA & C & -- & -- & -- & 90.0 & -- \\
Video Policy & C & -- & -- & -- & 94.0 & -- \\
CogVLA & C & 98.6 & 98.8 & 96.6 & 95.4 & 97.4 \\
\midrule
LingBot-VA & L & $98.5\pm0.3$ & $99.6\pm0.3$ & $97.2\pm0.2$ & $98.5\pm0.5$ & 98.5 \\
Cosmos Policy (Predict2) & C & 98.1 & 100.0 & 98.2 & 97.6 & 98.5 \\
Cosmos Policy (Predict2.5) & K & 98.7 & 99.7 & 98.3 & 96.6 & 98.33 \\
\midrule
\textbf{LingBot-VA + CBST} & Ours & 99.0 & 99.6 & 97.73 & 98.60 & 98.73 \\
\textbf{Predict2.5 + CBST} & Ours & 99.27 & 99.8 & 98.73 & 97.33 & 98.78 \\
\bottomrule
\end{tabular}}
\end{table}

\subsection{RoboCasa}
\label{app:robocasa_results}

Table~\ref{tab:robocasa_complete} includes all methods in the RoboCasa comparison reported by Cosmos Policy, together with its Predict2.5 update. Training demonstrations per task are retained because the methods use different data budgets. The CBST entry reports the robot data used by the underlying policy and the additional planner, with overlapping demonstrations counted only once.

\begin{table}[t]
\centering

\caption{Complete RoboCasa comparison over 24 tasks. Baseline scores and demonstration counts follow \citep{cosmospolicy2026}. Predict2.5 results follow the official Cosmos Cookbook \citep{cosmoscookbook}. Synthetic data are identified separately.}
\label{tab:robocasa_complete}
\begin{tabular}{@{}lcc@{}}
\toprule
Method & Demos per task & Average SR (\%) \\
\midrule
GR00T-N1 & 300 & 49.6 \\
UVA & 50 & 50.0 \\
DP-VLA & 3000 & 57.3 \\
GR00T-N1 + DreamGen & 300 (+10000 synthetic) & 57.6 \\
GR00T-N1 + DUST & 300 & 58.5 \\
UWM & 1000 & 60.8 \\
$\pi_0$ & 300 & 62.5 \\
GR00T-N1.5 & 300 & 64.1 \\
Video Policy & 300 & 66.0 \\
FLARE & 300 & 66.4 \\
GR00T-N1.5 + HAMLET & 300 & 66.4 \\
\midrule
Cosmos Policy (Predict2) & 50 & 67.1 \\
Cosmos Policy (Predict2.5) & 50 & 71.1 \\
\midrule
\textbf{Predict2.5 + CBST} & 50 & 72.44 \\
\bottomrule
\end{tabular}
\end{table}

\subsection{Additional Qualitative Examples}
\label{app:simulation_visualizations}
\begin{figure}[t]
    \centering
    \includegraphics[width=\linewidth]{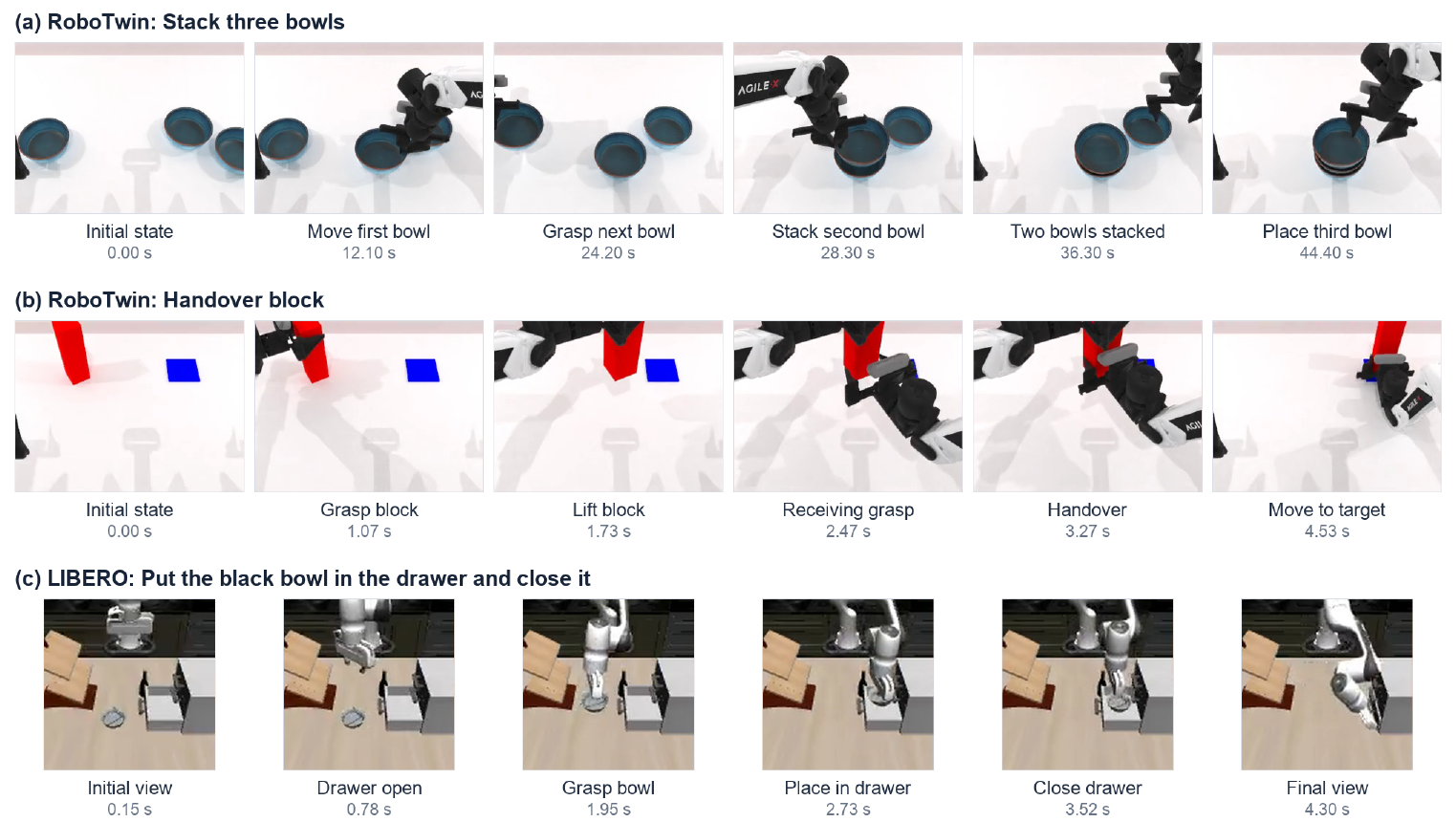}
    \caption{Additional recorded LingBot-VA executions:(a) stacking three bowls in RoboTwin, (b) handing over a block in RoboTwin, and (c) placing a black bowl in a drawer and closing it in LIBERO. Each row shows six temporally ordered frames;timestamps indicate video playback time.}
    \label{fig:simulation_more_rollouts}
\end{figure}

Figures~\ref{fig:simulation_more_rollouts} provide three additional simulation task sequences. Each example contains temporally ordered frames from a recorded execution. Task names and recording sources are specified in the captions. Annotations describe observable object interactions and state changes; model predictions are included only when corresponding outputs are available.

\section{Additional Ablation Studies}
\label{app:ablations}

This section specifies the ablation variants and extends the analysis in Section~\ref{sec:ablations}. We distinguish changes to pretraining, robot adaptation objectives, and inference-time conditioning. This distinction identifies which comparisons require independently trained planners and which can be evaluated using a shared checkpoint.

\subsection{Variant Definitions}
\label{app:ablation_variants}

\paragraph{Common configuration.}
Unless explicitly changed, variants use the same robot demonstration split, observation inputs, candidate variable domains, reference mechanisms, renderer, and frozen WAM checkpoint. Retrained variants use the same robot adaptation budget and checkpoint-selection rule. Inference-only variants share the full CBST checkpoint and differ solely in the information rendered for the WAM.

\begin{table}[t]
\centering

\caption{Definitions of the principal ablation variants. ``Retrain'' refers to robot-side planner adaptation; the WAM remains frozen in every variant.}
\label{tab:ablation_definitions}
\begin{tabularx}{\linewidth}{@{}p{0.27\linewidth}cX@{}}
\toprule
Variant & Retrain & Change relative to CBST \\
\midrule
Without human pretraining & Yes & Omit SSv2 pretraining while retaining the common visual initialization and robot adaptation procedure. \\
Shuffled human pretraining & Yes & Shuffle sampled frames within each human clip before pretraining; retain labels, frame content, and pretraining budget. \\
State-only & No & Render admissible goals and predicted semantic states without the selected interaction-dependency fields. \\
Full structure & No & Render all candidate parent fields instead of the contextually selected subset. \\
Random selection & No & Render randomly selected candidate parent fields, matching the learned subset cardinality. \\
Without mechanism consistency & Yes & Set $\lambda_m=0$ while preserving the remaining supervision and reference mechanisms. \\
\bottomrule
\end{tabularx}
\end{table}

\paragraph{Without human-video pretraining.}
This variant removes the human-video classification stage. The visual encoder retains the common initialization specified in Appendix~\ref{app:implementation}, and components normally initialized through human pretraining use their corresponding untrained initialization. Robot heads and selectors follow the same initialization procedure as full CBST. The comparison holds the robot adaptation budget fixed, rather than equating the total compute spent across human pretraining and adaptation.

\paragraph{Shuffled human-video pretraining.}
We randomly permute the sampled frame order within each human clip while keeping its original action label. The sampled images, class distribution, architecture, optimization settings, and number of pretraining updates are otherwise unchanged. Robot adaptation subsequently uses correctly ordered robot observations. This control evaluates the contribution of temporal ordering during human pretraining.

\paragraph{State-only conditioning.}
Using the full CBST checkpoint, we retain the original task instruction, admissible goals, and predicted semantic-state fields. Fields describing the selected parent dependencies and associated interval-interaction variables are omitted from the rendered condition. The underlying state predictions remain unchanged, so this comparison isolates the information presented to the WAM rather than differences in state-estimation training.

\paragraph{Full-structure conditioning.}
For every mechanism, we replace the selected parent set with the complete candidate parent set, $C_{i,t}=\mathrm{PA}_i$. Parent values are obtained from the same predictions as in CBST. Variable naming, field ordering, update frequency, and text encoding remain unchanged. This variant evaluates whether selecting a context-specific subset is useful beyond supplying all available structural fields.

\paragraph{Random selection.}
For mechanism $i$ at time $t$, let $k_{i,t}$ be the number of parents retained by CBST. We sample a subset uniformly from the candidate subsets of $\mathrm{PA}_i$ having cardinality $k_{i,t}$. The sampled subset uses the same predicted variable values and renderer as CBST. Sampling does not use ground-truth task outcomes or reference-validity labels. Random-selection seeds are fixed before evaluation and reported separately from training and rollout seeds.

\paragraph{Without mechanism consistency.}
We remove $\mathcal L_{\mathrm{mech}}$ from robot adaptation by setting $\lambda_m=0$. Semantic-state supervision, interval-variable supervision, and beam supervision remain unchanged. The reference mechanisms are still used to construct acceptable parent sets; the ablation therefore removes the mechanism-consistency training signal without removing the structural supervision source.

\subsection{Results across Backbones and Benchmarks}
\label{app:ablation_complete_results}

Table~\ref{tab:ablation_complete} extends the principal ablations to WAM backbones and associated benchmarks. RoboTwin Easy and Hard are reported separately. Published backbone scores are included as references and are not treated as paired measurements from the ablation runs.

\begin{table}[t]
\centering
\scriptsize
\caption{Principal ablation results across benchmarks. All values are success rates (\%). RoboTwin and LingBot-VA reference values follow \citep{lingbotva2026}; Predict2.5 reference values follow \citep{cosmoscookbook}.}
\label{tab:ablation_complete}
\begin{tabular}{@{}lccc@{}}
\toprule
& \multicolumn{3}{c}{LingBot-VA}\\
\cmidrule(lr){2-4}\
Variant & RoboTwin Easy & RoboTwin Hard & LIBERO-Long\\
\midrule
Backbone & 92.93 & 91.55 & $98.5\pm0.5$\\
\textbf{CBST} & 94.00 & 92.54 & 98.60 \\
\midrule
Without human pretraining & 92.80 & 91.38 & 98.00 \\
Shuffled human pretraining & 92.94 & 91.54 & 98.47 \\
State-only & 92.82 & 91.36 & 98.13 \\
Full structure & 93.42 & 91.96 & 98.53 \\
Random selection & 92.56 & 91.08 & 96.60 \\
Without mechanism consistency & 93.18 & 91.70 & 98.53 \\
\bottomrule
\end{tabular}
\end{table}



\subsection{Temporal Architecture}
\label{app:architecture_ablations}

We compare GRU, Transformer, temporal convolutional network (TCN), and MLP temporal modules while retaining the same visual encoder, observation samples, robot supervision, and WAM interface. The MLP receives the sampled frame features in their fixed temporal order through concatenation; the other modules process the corresponding feature sequence. Architecture-specific widths and depths, parameter counts, and optimization settings are reported with the results.

The comparison uses the same data and training schedule. Parameter counts are reported explicitly rather than assuming that the architectures have identical capacity.

\begin{table}[t]
\centering
\caption{Temporal architecture comparisons on LIBERO-Long. Parameter counts refer to the planner; latency refers to a planner update with the same input specification and accelerator.}
\label{tab:architecture_ablations}
\begin{tabular}{@{}lcccc@{}}
\toprule
Temporal module & Parameters & LingBot-VA SR & Predict2.5 SR & Planner latency \\
\midrule
MLP & 12.0 & 98.53 & 97.13 & 9 \\
TCN & 12.3 & 98.53 & 97.20 & 11 \\
GRU & 12.5 & 98.60 & 97.33 & 12 \\
Transformer & 13.2 & 98.66 & 97.53 & 13 \\
\bottomrule
\end{tabular}
\end{table}

\subsection{Training and Validation Curves}
\label{app:ablation_curves}

We report human-pretraining and robot-adaptation curves separately. Human-pretraining plots include classification loss and validation accuracy for correctly ordered and shuffled clips. Robot-adaptation plots separate semantic-state, interval-variable, mechanism-consistency, and structural-selection objectives, together with the corresponding held-out diagnostics. Curves use the original logged values. Training steps and epochs are identified explicitly, and resumed training segments retain their boundaries. Duplicate logging streams are not counted as additional optimization steps. Model selection follows the rule specified in Appendix~\ref{app:implementation}.

\subsection{Sensitivity to Observation History}
\label{app:history_sensitivity}
We vary observation history along two axes. First, we change the number of sampled frames while holding the lookback duration fixed, testing sampling density. Second, we change the lookback duration while holding the number of sampled frames fixed, testing temporal coverage. The current observation is retained in every setting. Other planner and WAM settings remain unchanged. Each sampling configuration is used consistently during adaptation and evaluation. The tested values and the default configuration are specified in Table~\ref{tab:sensitivity_configuration}.

\subsection{Sensitivity to Structural Compactness}
\label{app:compactness_sensitivity}
We vary the size-regularization coefficient $\lambda_{\mathrm{size}}$ while retaining the same semantic supervision, reference mechanisms, and acceptable-set construction. A separate planner is trained for each coefficient using the same adaptation budget. Setting $\lambda_{\mathrm{size}}=0$ removes the explicit compactness preference while preserving the structural objectives. For each setting, we report control success, the mean retained-parent fraction, and the number of text tokens delivered to the WAM. Token counts are measured using each backbone's native tokenizer. Structural compactness and prompt length are reported separately because equal parent counts need not produce equal token counts.

\begin{table}[t]
\centering

\caption{Sensitivity-analysis configurations. Default values correspond to the full CBST configuration.}
\label{tab:sensitivity_configuration}
\begin{tabularx}{\linewidth}{@{}p{0.25\linewidth}ccX@{}}
\toprule
Factor & values & Reference & Fixed quantities \\
\midrule
Sampled frame count
& $\{4,8,16,32\}$
& $16$
& Lookback duration and temporal sampling strategy. \\
Lookback duration (s)
& $\{1,2,4,8\}$
& $4$
& Frame count and recent/history allocation ratio. \\
$\lambda_{\mathrm{size}}$
& $\{0,10^{-3},10^{-2},10^{-1}\}$
& $10^{-2}$
& Other loss weights, reference mechanisms, and training budget. \\
\bottomrule
\end{tabularx}
\end{table}

\subsection{Computational Cost}
\label{app:ablation_cost}

We measure the cost of the additional planner and conditioning path separately from WAM generation. Measurements use a fixed accelerator, input specification, batch size, and action-generation configuration for each backbone. Warm-up iterations are excluded, and accelerator operations are synchronized at the declared timing boundaries.

Planner time includes visual-temporal encoding, semantic prediction, and parent-set selection. Conditioning time separates text rendering from native text encoding. End-to-end request latency is measured directly rather than inferred by summing components that may overlap. Simulator stepping and video recording are excluded from policy-request timing.

\section{Real-World Experimental Details}
\label{app:realworld_details}

\subsection{Platform and Task Formulation}
The platform consists of two AgileX manipulators with parallel-jaw grippers operating over a shared tabletop. Each task is specified by a manipulated object and a fixed target region or receptacle. The controller must acquire the object, transport it, and establish the required final placement. When a single arm cannot reach both the source and destination, the arms coordinate through their overlapping reachable workspace. No particular handover trajectory is prescribed by the task. The supplied videos show a fixed external recording view.

\subsection{Task Definitions}
\label{app:realworld_tasks}
Table~\ref{tab:realworld_tasks} describes the ten task entries. All names specify placement outcomes rather than a required handover.Target locations are fixed within a task configuration; any changes across configurations must be documented separately from variation in initial object pose. The detailed results in Table~\ref{tab:realworld_detailed}.

\begin{table}[t]
\centering

\setlength{\tabcolsep}{4pt}
\caption{Task inventory. }
\label{tab:realworld_tasks}
\begin{tabularx}{\linewidth}{@{}lp{0.24\linewidth}>{\raggedright\arraybackslash}X@{}}
\toprule
ID & Task & Placement objective and success condition \\
\midrule
T1 & Remote-control placement & Move the remote from the tabletop into the designated open container; it must remain fully inside after gripper release. \\
T2 & Cup placement & Move the cup to the designated tabletop region and leave it stable after release; handover may be used to connect the source and target workspaces. \\
T3 & Plate racking & Place the plate into the designated rack slot, with the rack supporting the plate after both grippers disengage. \\
T4 & Apple placement & Place an apple inside the designated fruit basket and release it. \\
T5 & Banana placement & Place a banana within the designated tray and release it. \\
T6 & Orange placement & Place an orange inside the designated bowl and release it. \\
T7 & Bottle placement & Place a bottle upright within the designated tabletop region and release it. \\
T8 & Sponge placement & Place a sponge inside the designated storage box and release it. \\
T9 & Spoon placement & Place a spoon within the designated cutlery tray and release it. \\
T10 & Block placement & Place a block inside the designated box and release it. \\
\bottomrule
\end{tabularx}
\end{table}

\subsection{Baseline and Evaluation Protocol}
\label{app:realworld_protocol}
The comparison pairs Cosmos Policy with a Predict2.5 backbone and its CBST-augmented counterpart. The intended controlled comparison holds the task-adapted low-level policy checkpoint, robot observations, control settings, and evaluation conditions fixed, while adding CBST's contextual structural conditioning. The official Cosmos Policy ALOHA experiments provide a related tabletop-manipulation reference~\citep{cosmospolicy2026}; those published real-world results use Predict2, and the official Cookbook states that Predict2.5 was not trained on ALOHA~\citep{cosmoscookbook}. Those results are therefore not substituted for evaluation on our AgileX tasks.

Each method is evaluated for 20 trials on each of ten tasks. This specifies the number of executions, not the number of distinct layouts or random seeds. Initial object poses, target geometry, and the mapping of initializations between methods must be documented to establish comparability. For the final protocol, the same initialization set should be used for both methods, with scene resets between executions and balanced method order.

\paragraph{Success criteria.} Full-task success requires the designated object to occupy its specified target region or receptacle, satisfy the task-specific placement requirement, and remain stable after gripper release within the evaluation time limit. Handover is neither an additional task objective nor an independently rewarded outcome. Grasping or transferring an object without completing its final placement does not constitute success. The handling of autonomous retries, intermediate drops, and human interventions must follow the same documented rule for both methods.

\paragraph{Metrics.} Let $S_{m,k}$ denote the number of successful trials for method $m$ on task $k$. With 20 trials per task, the task success rate and ten-task average are
\begin{equation}
\mathrm{SR}_{m,k}=100\frac{S_{m,k}}{20},\qquad \overline{\mathrm{SR}}_m=\frac{1}{10}\sum_{k=1}^{10}\mathrm{SR}_{m,k}=100\frac{\sum_{k=1}^{10}S_{m,k}}{200}.
\end{equation}
Single-task rates therefore change in increments of 5 percentage points, while the ten-task average changes in increments of 0.5 percentage points. Improvements are reported as percentage-point differences. A mean computed over the three visualized tasks is explicitly labeled as a subset mean and is not used as the ten-task result. Stage-level annotations, if available, distinguish acquisition, transport, handover where applicable, and final placement; they are diagnostic and do not replace the full-task metric.

\begin{table}[t]
\centering

\setlength{\tabcolsep}{4pt}
\caption{Detailed real-world results. Each method has 20 trials per task. Gains are measured in percentage points.}
\label{tab:realworld_detailed}
\begin{tabular}{@{}llccccc@{}}
\toprule
 & & \multicolumn{2}{c}{Cosmos Policy (Predict2.5)} & \multicolumn{2}{c}{+ CBST} & \\
\cmidrule(lr){3-4}\cmidrule(lr){5-6}
ID & Task & Successes & SR (\%) & Successes & SR (\%) & Gain \\
\midrule
T1 & Remote & $(17/20)$ & $85$ & $(18/20)$ & $90$ & $5$ \\
T2 & Cup & $(14/20)$ & $70$ & $(17/20)$ & $85$ & $15$ \\
T3 & Plate & $(13/20)$ & $65$ & $(15/20)$ & $75$ & $10$ \\
T4 & Apple & $(10/20)$ & $50$ & $(12/20)$ & $60$ & $10$ \\
T5 & Banana & $(15/20)$ & $75$ & $(16/20)$ & $80$ & $5$ \\
T6 & Orange & $(8/20)$ & $40$ & $(9/20)$ & $45$ & $5$ \\
T7 & Bottle & $(17/20)$ & $85$ & $(17/20)$ & $85$ & $0$ \\
T8 & Sponge & $(16/20)$ & $80$ & $(17/20)$ & $85$ & $5$ \\
T9 & Spoon & $(11/20)$ & $55$ & $(14/20)$ & $70$ & $15$ \\
T10 & Block & $(16/20)$ & $80$ & $(18/20)$ & $90$ & $10$ \\
\bottomrule
\end{tabular}
\end{table}

\subsection{Qualitative Executions}
\label{app:realworld_visualizations}
\begin{figure}[t]
    \centering
    \includegraphics[width=\linewidth]{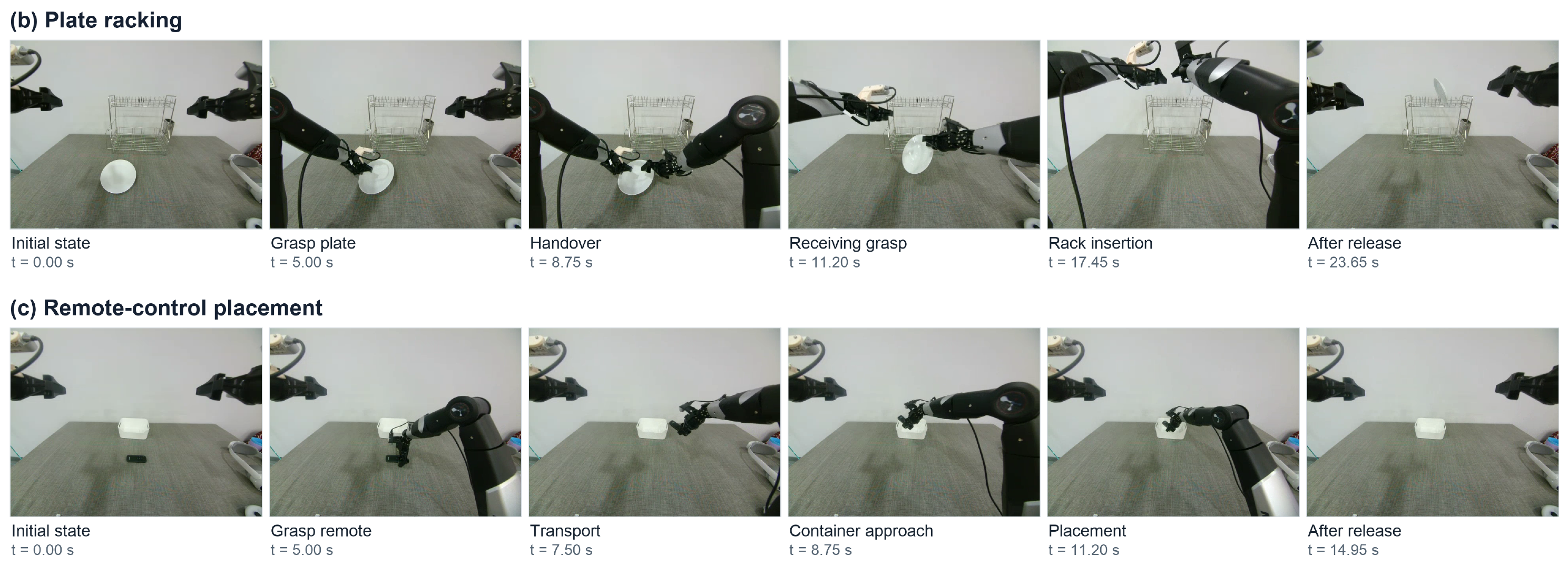}
    \caption{Additional real-world executions on the AgileX platform:
    (b) plate racking and (c) remote-control placement.
    Frames are shown in temporal order, with timestamps indicating
    video playback time. Annotations describe observed manipulation events.}
    \label{fig:realworld_additional}
\end{figure}

The three supplied recordings are identified by the authors as CBST executions. Each visualization contains six original video frames selected around visible manipulation events. In the cup-placement sequence shown in the main text, the initiating arm lifts the cup and transfers it to the receiving arm for final tabletop placement. Figure~\ref{fig:realworld_additional} shows plate transport into a rack and remote-control placement into a container. In the plate sequence, grasp transfer enables transport between arm workspaces; in the remote sequence, one arm completes the placement while the other remains inactive. These examples illustrate execution strategies and do not measure their frequency across the ten-task suite.

The figure labels describe visible events rather than predicted states or causal dependencies. All selected frames preserve the source image content. The encoded timestamps and frame rate are used only to identify frames, not to infer control frequency or execution latency. Selected successful videos provide qualitative examples and are not used to estimate task success rates.

\end{document}